\pdfoutput=1
\documentclass{article}

\usepackage{arxiv}

\usepackage[utf8]{inputenc}
\usepackage[T1]{fontenc}
\usepackage{microtype}
\usepackage{amsmath,amssymb,amsfonts}
\usepackage{mathabx}
\usepackage{amsthm}
\usepackage{mathrsfs}
\usepackage{bbm}
\usepackage{graphicx}
\usepackage{siunitx}
\usepackage[title]{appendix}
\usepackage{xcolor}
\usepackage{textcomp}
\usepackage{manyfoot}
\usepackage{booktabs}
\usepackage{multirow}
\usepackage{algorithm}
\usepackage{algorithmicx}
\usepackage{algpseudocode}
\usepackage{listings}
\usepackage{tabularray}
\usepackage{tabularx, ragged2e}
\usepackage{longtable}
\usepackage{pdflscape}
\usepackage{changepage}
\usepackage{pgf-pie}
\usepackage{wheelchart}
\usepackage{caption}
\usepackage{subcaption}
\usepackage{comment}
\usepackage{setspace}
\usepackage[normalem]{ulem}
\useunder{\uline}{\ul}{}

\usepackage{natbib}
\usepackage{hyperref}
\hypersetup{colorlinks=true,linkcolor=blue!60!black,citecolor=blue!60!black,urlcolor=blue!60!black}
\usepackage{xurl}
\usepackage{cleveref}

\DeclareUnicodeCharacter{202F}{\,}

\providecolor{OliveGreen}{cmyk}{0.64,0,0.95,0.40}
\DeclareRobustCommand{\gc}[1]{#1}

\title{Advancing Image-Based Grapevine Variety Classification with a New Benchmark and Evaluation of Masked Autoencoders}

\renewcommand{\shorttitle}{Grapevine Variety Classification with Masked Autoencoders}

\author{%
  \textbf{Gabriel A. Carneiro}$^{1,2}$\thanks{Corresponding author: \texttt{gabrielc@utad.pt}} \quad
  \textbf{Thierry J. Aubry}$^{3}$ \quad
  \textbf{Ant\'onio Cunha}$^{1,2}$ \quad
  \textbf{Petia Radeva}$^{4}$ \quad
  \textbf{Joaquim J. Sousa}$^{1,2}$ \\[2ex]
  \normalfont\small $^{1}$University of Tr\'as-os-Montes e Alto Douro, Vila Real, Portugal \\
  \normalfont\small $^{2}$INESC Technology and Science (INESC TEC), Porto, Portugal \\
  \normalfont\small $^{3}$C\^oa Parque, Funda\c{c}\~ao para a Salvaguarda e Valoriza\c{c}\~ao do Vale do C\^oa, Vila Nova de Foz C\^oa, Portugal
  \\ \normalfont\small $^{4}$Departament de Matem\`atiques i Inform\`atica, Universitat de Barcelona, Barcelona, Spain
}

\date{\gc{\footnotesize This work has been submitted to the IEEE for possible publication. Copyright may be transferred without notice, after which this version may no longer be accessible.}}

\begin{document}

\maketitle

\begin{abstract}
The identification of grapevine varieties from field images is a challenge, due to the subjectivity of ampelography, the cost of molecular analysis and the limited generalisation of deep learning models trained with transfer learning from natural-image datasets. This study evaluates Masked Autoencoders (MAE) as a self-supervised alternative capable of learning domain-specific representations directly from unlabelled vineyard images. A public benchmark of 43 grapevine varieties acquired in the field over five growing seasons is presented, divided into a multi-year set (Dataset 1) and a separate 2024 set (Dataset 2), and it is used to evaluate MAE pre-training on unlabelled grapevine images. On Dataset 1, a ViT-B/16 pre-trained with MAE on 54,571 unlabelled grapevine images reached a macro F1 score of 0.7776, against 0.7494 for the same architecture initialised from the public MAE ImageNet-1K (IN1K) checkpoint with the same fine-tuning, and 0.6817 for the best supervised IN1K baseline. Domain-specific pre-training benefited from long training and random cropping, outperformed IN1K initialisation for all fractions of labelled data and was only slightly affected by the mask ratio. However, the generalisation to a new season remained the main limitation, since the models trained on Dataset 1 reached F1 scores between 0.4120 and 0.4747 on the 2024 images, regardless of the initialisation, with the worst results in the early phenological stages. \gc{A feature-space analysis showed that each acquisition season is distinct, including the seasons used for training, and that the 2024 images of the early phenological stages are the most distant from the training data.} When the 2024 images were added to the training data, the F1 score on Dataset 1 increased to 0.7956. Therefore, domain-specific MAE pre-training is a competitive and label-efficient initialisation, but it does not replace labelled data from several seasons.

\end{abstract}

\keywords{Self-supervised learning \And Masked autoencoders \And Vision transformers \And Grapevine variety identification \And Precision viticulture}

\section{Introduction}
\label{sec:introduction}
Grapevine is one of the most widely cultivated and economically significant horticultural crops worldwide \citep{eyduran_sugars_2015}. Its importance stems from their use in wine and juice production, as well as the consumption of fruits and leaves, making them crucial to the economies of various countries  \citep{nascimento_early_2019}. It is estimated that between 5,000 and 8,000 grapevine varieties exist under approximately 24,000 different names  \citep{schneider_verifying_2001, cunha_portuguese_2009}. Identifying these varieties is essential for managing quality control and regulatory compliance in the wine production chain, as they influence acidity, sugar content, colour, taste and health properties of the product \citep{Rankine2017lnfluenceOG, Fanzone2012PhenolicCO}. Additionally, identifying and characterising local autochthonous varieties helps mitigate climate change effects by promoting varieties that adapt better to warmer, drier conditions, while also preserving indigenous varieties from extinction \citep{SanchoGaln2020IdentificationAC, robinson_wine_2013}.

Today, grapevine varieties are primarily identified by ampelography or molecular analysis  \citep{carneiro_deep_2024}. Ampelography, which relies on expert-led visual analysis of leaves and fruits, is inherently subjective and depends on professional expertise \citep{ampelografia, Pavek2003}. This approach faces additional challenges due to a scarcity of specialists \citep{P2004}. Molecular analysis, on the other hand, uses molecular markers through techniques like random amplified polymorphic DNA and amplified fragment length polymorphism \citep{P2004}. While it provides a solution for the subjectivity of ampelography, it is costly and time-intensive.

In recent years, computer vision techniques have emerged as cost-effective and rapid alternatives to overcome the limitations of ampelography and molecular analysis. Previous studies have explored machine learning classifiers applied to RGB and spectral images to classify grapevine varieties using samples of fruits, leaves, and seeds \citep{gutierrez_--go_2018, abbasi_data_2024, landa_accurate_2021}. With the advent of deep learning (DL), recent research has adopted DL-based classifiers and object detectors for the same purpose, using RGB \citep{de_nart_vine_2024, terzi_automatic_2024, magalhaes_toward_2023, liu_development_2021, carneiro_evaluating_2023, carneiro_2023, carneiro_grapevine_2022}, hyperspectral \citep{lopez_classification_2024}, and spectral data \citep{FERNANDES2019104855} acquired from leaves and fruits.

Convolutional Neural Networks (CNNs) have been widely explored for grapevine classification using RGB images. For instance, \citet{de_nart_vine_2024} classified 27 different grapevine varieties using images of leaves acquired in the field and in a controlled environment, testing models such as MobileNetV2 \citep{sandler_mobilenetv2_2018}, EfficientNet \citep{tan_efficientnet_2019}, ResNet \citep{resnet}, InceptionResNetV2 \citep{DBLP:journals/corr/SzegedyIV16}, and Inception V3 \citep{szegedy_rethinking_2015}. Inception V3 achieved an accuracy of 99.70\%, but a significant drop in performance was observed when tested on an external dataset \citep{Vlah2021}. In this dataset, the best result was achieved by EfficientNet, with an accuracy of 35.00\%. Similarly,  \citet{terzi_automatic_2024} classified between 10 and 11  varieties using images of fruits and leaves acquired in the field. The authors tested a handcrafted model, GoogLeNet \citep{szegedy_going_2014}, and AlexNet \citep{NIPS2012_c399862d}. The best performance was achieved by the handcrafted model with an overall accuracy of 97.20\%. Moreover, the results obtained indicate that identifying grapevine varieties using fruit images led to slightly better results than classifying them using leaf images.  \citet{carneiro_2023} examined the impact of background information for the identification of 12 varieties, showing that removing secondary information slightly decreased model performance.

In addition to CNNs, Transformer-based architectures, such as Vision Transformers (ViTs) \citep{dosovitskiy_image_2020}, have gained prominence.  \citet{carneiro_grapevine_2022} and \citet{kunduracioglu_advancements_2024} applied ViTs and Swin-Transformers to classify grapevine varieties, achieving superior performance compared to CNNs, especially with images obtained in controlled environments.

Although these studies have demonstrated high performance in classifying grapevine varieties, the data used does not accurately reflect real-world conditions when applying these classifiers in the field. This limitation arises from a lack of representativeness, which could lead to poor performance \citep{carneiro_deep_2024}. The primary drawback of these studies is the limited number of grapevine varieties and growing seasons included in the datasets. Additionally, data availability remains a challenge. Studies that analysed more than 20 varieties \citep{de_nart_vine_2024, magalhaes_toward_2023, terzi_automatic_2024} did not use publicly available benchmarks. In contrast, freely available datasets typically include fewer than 10 varieties \citep{koklu_cnn-svm_2022, al-khazraji_image_2023, santos_embrapa_2019, sozzi_wgrapeunipd-dl_2022}, while those covering a greater number of varieties contain fewer than 2,500 samples \citep{Vlah2021, seng_computer_2018}. Moreover, all of these datasets were collected during a single growing season, further limiting their generalisability and, consequently, the applicability of models trained on them in real-world scenarios. Therefore, a dataset encompassing multiple grapevine varieties, collected across different locations and growing seasons, is essential for effectively evaluating DL-based methods with the aim of classifying grapevine varieties.

Another drawback is that all studies based on DL and RGB images heavily rely on transfer learning from IN1K \citep{deng2009imagenet}, a collection of natural images that substantially differs from images of grapevine leaves or fruits, leading to domain shift and supervision collapse in models \citep{el-nouby_are_2021, doersch_crosstransformers_2021}. There is a significant difference between IN1K, composed of general scene images, and images of centred leaves or fruits acquired in the field, which results in a large domain shift between the source dataset and the transfer learning target dataset. On the other hand, supervision collapse refers to the fact that the model learns to map images and labels during the supervised pre-training phase, which can lead to the elimination of relevant information for the target tasks.

Considering these concerns and the high cost of annotating datasets, pre-training methods have been developed with focus on learning without labelled data. These methods, which aim to improve label efficiency, include unsupervised learning, self-supervised learning, and weak learning \citep{9442775, LI2023108412}. Focussing on SSL, this family of methods are able to learn directly from data without relying on external labels, serving as effective pre-training strategies \citep{el-nouby_are_2021}. The main goal of SSL is to generate representations that can be used in a downstream task, and it is usually based on joint-embedding architectures, where two networks are trained to produce similar representations for different views of the same image \citep{bardes_vicreg_2022}. The main challenge related to SSL approaches is the generation of \textit{collapsed} representations, a constant output vector for any input \citep{balestriero_cookbook_2023}. To avoid collapse, different methods use different strategies, such as large batch sizes \citep{chen_simple_2020}, memory banks \citep{he_momentum_2020}, asymmetric architectures \citep{chen_exploring_2020}, clustering \citep{caron_unsupervised_2021}, or maximising information \citep{bardes_vicreg_2022}. Unlike these, other methods are based on image reconstruction \citep{he_masked_2021} using an encoder-decoder architecture.

The large number of publicly available images and the difference between agricultural and general scene images made SSL methods a good alternative for IN1K knowledge transfer in the agricultural context.  Recently, several studies have compared the use of SSL methods based on joint-embedding architectures and the transfer of knowledge from IN1K using agricultural images.  \citet{guldenring_self-supervised_2021} applied SwAV \citep{caron_unsupervised_2021} to evaluate the application of SSL to agricultural images. ResNet-34 was pre-trained using different configurations and then the representations were used for plant classification and segmentation. The best performance was obtained by models pre-trained with SwAV, initialised with IN1K weights, achieving an accuracy of 94.90\%.  \citet{kar_self-supervised_2023} compared the performance of NNCLR \citep{dwibedi_little_2021}, BYOL \citep{grill_bootstrap_2020} and BarlowTwins \citep{zbontar_barlow_2021} with IN1K initialisation to perform pest classification using ResNet-18 and ResNet-50. The best performance was obtained by ResNet-50 initialised with BYOL pre-training weights, with an accuracy of 94.16\%.
Obtaining different results, \citet{ogidi_benchmarking_2023} compared the performance of supervised pre-training with IN1K, DenseCL \citep{wang_dense_2021} and MocoV2 \citep{chen_improved_2020} on wheat head detection, plant instance detection, wheat spikelet counting and leaf counting. IN1K knowledge transfer performed better than SSL initialisation in all tasks except leaf counting. The authors also presented an analysis with redundant data, a common characteristic in agricultural datasets acquired with vehicles. The models initialised with SSL methods were more sensitive and performed worse in this scenario.

The Masked Autoencoders method \citep{he_masked_2021} (MAE), based on image reconstruction, was also explored. Unlike other SSL methods based on joint-embedding architectures, MAEs do not require large batch sizes, strong data augmentation, or architectural asymmetries to prevent feature collapse \citep{he_masked_2021}. Because they do not rely on batch statistics, MAEs are expected to be less sensitive to data imbalance and to be more efficient in terms of labels, since MAE can be seen as a denoising autoencoder, and, unlike contrastive-based methods, it preserves local relationships in representations, a property that is potentially useful for fine-grained tasks such as grapevine variety classification \citep{assran_hidden_2022, el-nouby_are_2021,park_what_2023}. In addition, \citet{el-nouby_are_2021} showed that autoencoder pre-training remains competitive when only small datasets are available. This is important in the agricultural context, since large labelled datasets are rare and the available domain-specific unlabelled datasets are much smaller than IN1K. Therefore, the present study evaluates how this approach behaves in a fine-grained, multi-season agricultural problem.

MAE was applied in the agricultural context for the identification of oxidation \citep{kang_identification_2023}, flower detection \citep{li_research_2024}, and leaf disease classification \citep{wang_classification_2024, liu_self-supervised_2022}. \citet{kang_identification_2023} combined MAE and supervised learning with the aim of identifying different levels of oxidation in nuts. The proposed approach outperformed IN1K initialisation, achieving an F1 accuracy score of 94.80.  \citet{li_research_2024} used MAE to perform flower detection, training a ViT-L to be used as the backbone of a YOLOv5 \citep{Jocher_YOLOv5_by_Ultralytics_2020}, obtaining a mAP of 71.30.  \citet{wang_classification_2024} used MAE to train a modified version of ViT to classify leaf diseases, achieving an accuracy of 99.35\%. Finally, \citet{liu_self-supervised_2022} applied MAE to pre-train a ViT to classify leaf pests and diseases. The authors replaced random masking with a method to select the patches that would be masked, considering the similarity between the patches within the image. The modification increased the model's performance for three downstream tasks tested. Despite these applications achieving remarkable results, there are factors that still need to be tested when using MAE as a pre-training tool in the agricultural context. All of these studies, except \citet{liu_self-supervised_2022}, used the default MAE configuration for training IN1K, therefore no ablation studies were presented regarding low-data regime training, long pre-training and the modification of training hyperparameters such as learning rate, mask ratio and data augmentation. Moreover, to the best of our knowledge, SSL pre-training has not yet been systematically evaluated for grapevine variety identification from field-acquired leaf images.

Given these factors, this study presents an evaluation of MAEs for grapevine variety identification using field-acquired images over different seasons. The main contributions of this study are the following.

\begin{enumerate}
	\item A public benchmark for the classification of 43 grapevine varieties from field-acquired leaf images, collected over five growing seasons (2020 to 2024) and in different field conditions, which is, to our knowledge, the largest public benchmark for this task in number of varieties, with a separate 2024 set (Dataset 2) for cross-season testing;

	\item An evaluation of MAE pre-training on unlabelled grapevine images, including ablations on the pre-training data, initialisation, training duration, data augmentation, mask ratio and amount of labelled data;

	\item A preliminary comparison, with limited computational resources, between MAE and two other SSL methods (DINO and Barlow Twins);

	\item An analysis of the performance on a new growing season, including week-by-week results across phenological stages and a calibrated quantification of the distribution shift between seasons.

\end{enumerate}

The methodology involves pre-training ViT models with MAEs, followed by transfer learning and fine-tuning of the resulting weights. Models' performance was assessed quantitatively using metrics such as Precision, Recall, F1-Score and feature similarity through Linear Centered Kernel Alignment, and qualitatively using Attention Maps. The seasonal analysis was conducted using domain shift quantification through Maximum Mean Discrepancy and a classifier-based discriminator. Furthermore, a comparison between MAE and other SSL methods, namely DINO \citep{caron_emerging_2021} and Barlow Twins \citep{zbontar_barlow_2021}, was performed to provide practical insights into the different approaches.

The results show that domain-specific MAE pre-training is a competitive and label-efficient initialisation for grapevine variety identification, but they also show that models trained on one set of seasons lose much of their accuracy on a new season, regardless of how they were initialised. By releasing a multi-season benchmark and reporting both outcomes, this study aims to provide a realistic baseline for future work on reliable vineyard applications such as variety monitoring, conservation of local varieties and quality control in grape and wine production.

\section{Material and Methods}
\label{sec:methods}
\subsection{Datasets}

This study introduces two labelled datasets designed to evaluate the efficacy of DL-based models in classifying grapevine varieties along with two additional unlabelled datasets. An overview of them with their main characteristics can be seen in \gc{\Cref{tab:grapevine_datasets}}. The datasets include images of grapevine leaves collected over several seasons and under diverse environmental conditions. By capturing various grapevine varieties, leaf health conditions, and phenological stages, the labelled datasets aim to address the need for representativeness in model training and evaluation. Each dataset serves a distinct purpose. The labelled datasets (Dataset 1 and Dataset 2) are used for model training and validation in the classification task, while two unlabelled datasets (Dataset 3 and its extension, Dataset 3+) are used for MAE pre-training in a self-supervised context. This approach ensures a comprehensive assessment of MAEs across different data regimes and conditions, thereby enhancing their applicability to precision viticulture.

\begin{table}
	\centering
	\caption{Summary of the used datasets.}
	\label{tab:grapevine_datasets}
	\setlength{\tabcolsep}{5pt} %
	\begin{tabularx}{\linewidth}{@{}lXXXX@{}}
		\toprule
		\textbf{Category} & \textbf{Dataset 1} & \textbf{Dataset 2} & \textbf{Dataset 3} & \textbf{Dataset 3+} \\
		\midrule
		Dataset Purpose & Baseline dataset capturing grapevine leaf images over multiple years & Temporal extension capturing grapevine leaf images during the 2024 growing season & Unlabelled dataset for SSL pre-training & Extended version of Dataset 3 with additional images for diversity \\
		\addlinespace
		Location of Acquisition & Vila Real, Portugal & Vila Real, Portugal & Various locations & Various locations \\
		\addlinespace
		Time Period & 2020 to 2023 & 2024 & Various & Various \\
		\addlinespace
		Acquisition Devices & Smartphones, Tablets, Cameras & Smartphones & Various sources, mixed acquisition devices & Various sources, mixed acquisition devices \\
		\addlinespace
		Image Distance & 20 to 40 cm from plant & 20 to 40 cm from plant & Various distances & Various distances \\
		\addlinespace
		Dataset Size & 7,939 (training), 1,508 (validation) and 4,032 (test) samples & 4,574 (training), 1,060 (validation) and 1,142 (test) samples & 33,671 images & 54,571 images \\
		\addlinespace
		Number of Varieties & 43 Classes & 43 Classes & Unlabelled (used for SSL pre-training) & Unlabelled (used for SSL pre-training) \\
		\addlinespace
		Publicly Available & Yes & Yes & No & No \\
		\bottomrule
	\end{tabularx}
\end{table}

\subsubsection{UTAD Grapevine Varieties Dataset 2023 (Dataset 1)}\label{sec:dataset1}

\begin{figure}
	\center
	\includegraphics[width=0.8\textwidth]{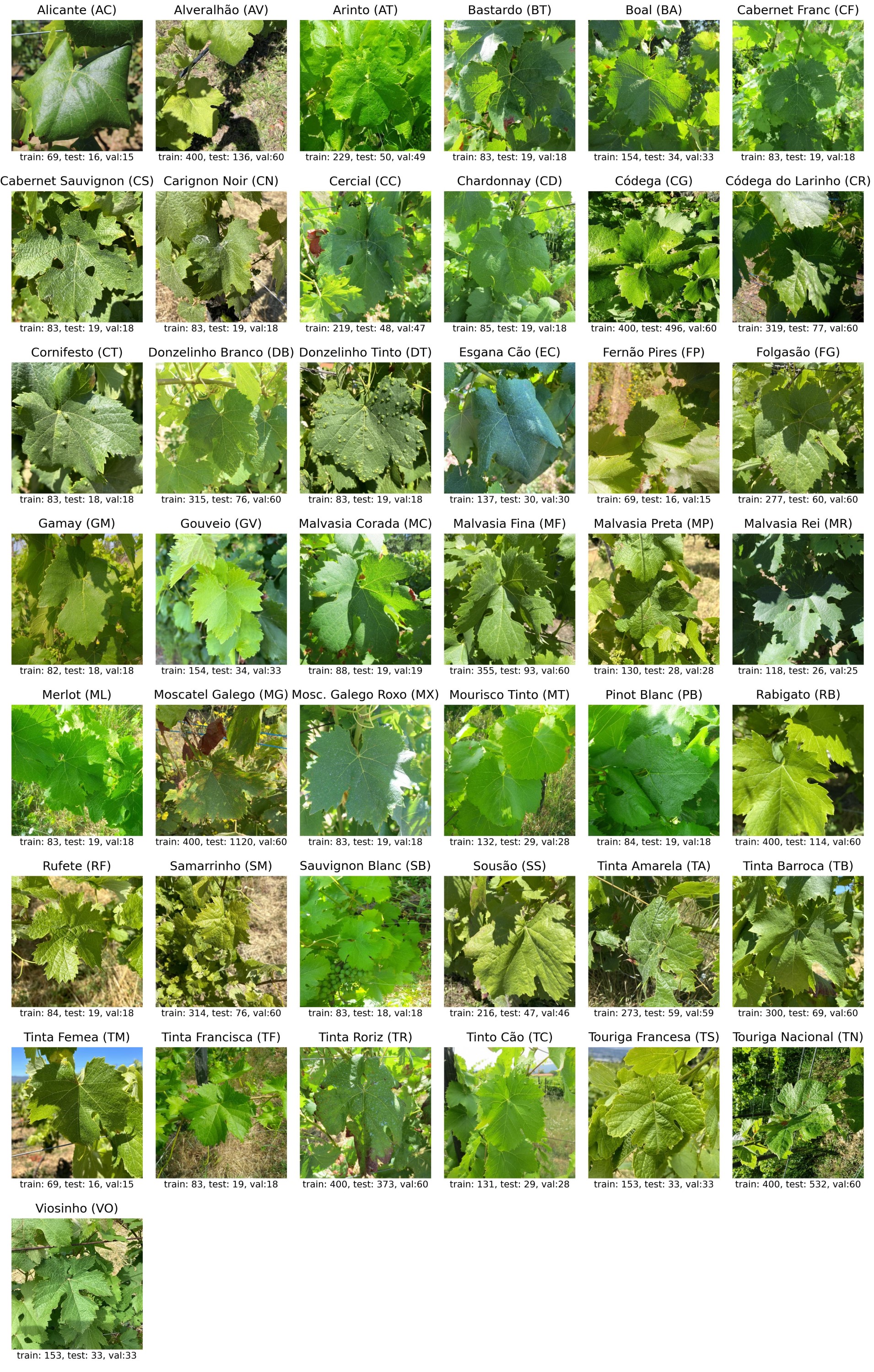}
	\caption{Examples for each class in Dataset 1. Below each image is the number of images for the class in each subset of the dataset. The two-letter codes in brackets are used to identify the classes in the tables; the confusion matrices in \Cref{fig:cms} use the same alphabetical order, numbered from 0 to 42. \label{fig:dataset1_samples}}
\end{figure}

Dataset 1 was created through the acquisition of grapevine leaf images in natural settings over four years (2020 to 2023) at two sites in Vila Real, Portugal, namely the University of Trás-os-Montes and Alto Douro (UTAD) and Palácio de Mateus. \Cref{fig:dataset1_samples} shows samples of each class. This dataset includes 35 grapevine varieties grown at UTAD and 26 at Palácio de Mateus, capturing five of the thirteen most commonly cultivated grape varieties worldwide (Cabernet Sauvignon, Merlot, Tinta Roriz, Chardonnay, and Pinot Noir)   \citep{noauthor_distribution_nodate}. Additionally, it includes major varieties used in Douro Demarcated Region wine production (such as Touriga Nacional, Touriga Franca, and Tinta Roriz) and 26 of the 29 varieties recommended for Port wine production, with only three missing (Síria, Castelão, and Verdelho)  \citep{noauthor_dop_nodate2}.

The images were acquired using smartphones (Apple iPhone 11, 12, and 14, and Blackberry Z10), tablets (Samsung Galaxy Tab S6 Lite and Microsoft Surface Go 2) and cameras (Canon EOS 600D, Canon EOS T100 and Canon IXUS 127) at a distance of between 20 and 40 cm from the plant. At least 10 different plants were used to acquire data for each variety. In terms of timing, the images were acquired within the season on different dates (between May and August), ensuring that different phenological stages were represented. It is worth noting that most of the images were acquired in 2023. The images included healthy and unhealthy leaves, in different positions on the grapevines, and plants of different ages. In addition, daylight variations was also included in the dataset. Considering that the dataset was contributed by different people, there was no pattern to the number of samples acquired per day or class. Capturing images across different devices and lighting conditions was intentional, as real-world viticulture deployments operate under heterogeneous acquisition scenarios. Including such diversity reduces the risk of overfitting to a single imaging condition.

\begin{figure}
	\center
	\includegraphics[width=\textwidth]{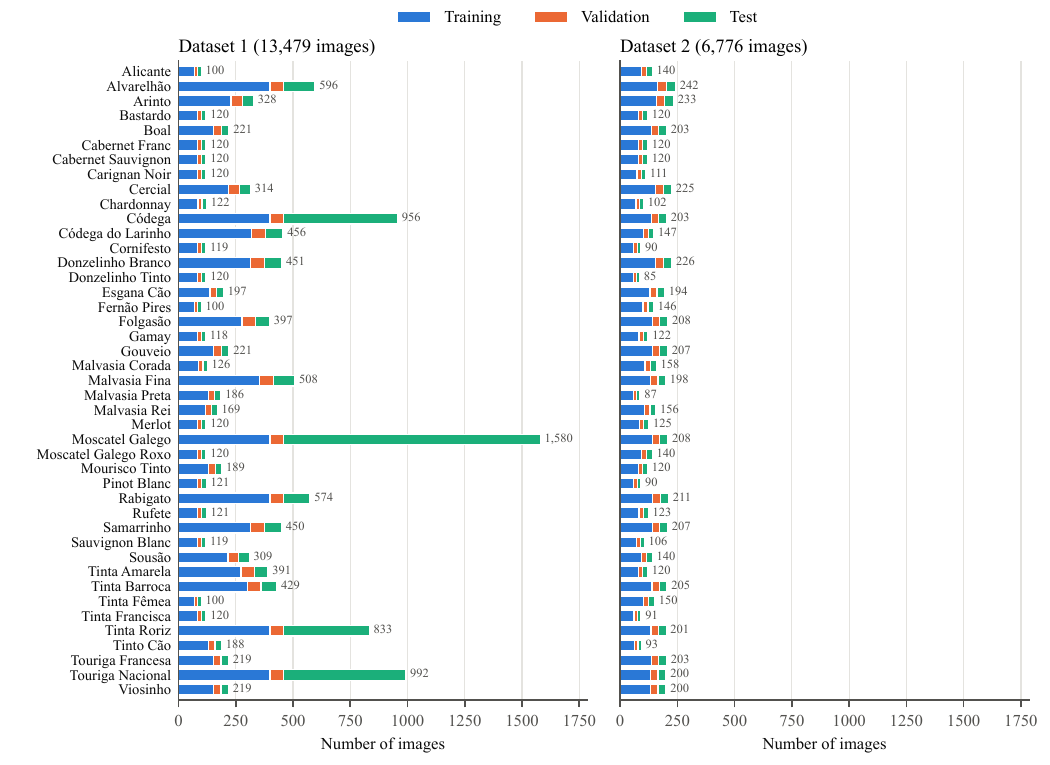}
	\caption{Composition of Dataset 1 (left) and Dataset 2 (right) by class, with the training, validation and test images of each class stacked and the total number of images at the end of each bar. \gc{Both panels use the same horizontal axis, so the bar lengths can be compared between the two datasets.} \label{fig:ds_composition}}
\end{figure}

Following acquisition, the dataset was split into training, validation and test subsets. \gc{The images were assigned to the subsets at random, so images of the same plant can appear in more than one subset.} Due to the large imbalance between the classes, the number of images per class was limited to 400 in the training subset and to 60 in the validation subset, so that the models are trained with a more balanced dataset. The remaining images were assigned to the test subset, which is therefore imbalanced (16 to 1,120 images per class, see \Cref{fig:dataset1_samples}), and, for this reason, all the scores are reported as macro averages over the classes. In total, the dataset comprises 7,939 samples for training, 1,508 for validation, and 4,032 for testing. The number of samples per class can be seen in \Cref{fig:dataset1_samples}, and the composition of the subsets can be seen in \Cref{fig:ds_composition}. Dataset 1 is publicly available at \url{https://zenodo.org/records/15160944}.

\subsubsection{UTAD Grapevine Varieties Dataset 2024 (Dataset 2)}

\begin{figure}
	\center
	\includegraphics[width=0.5\textwidth]{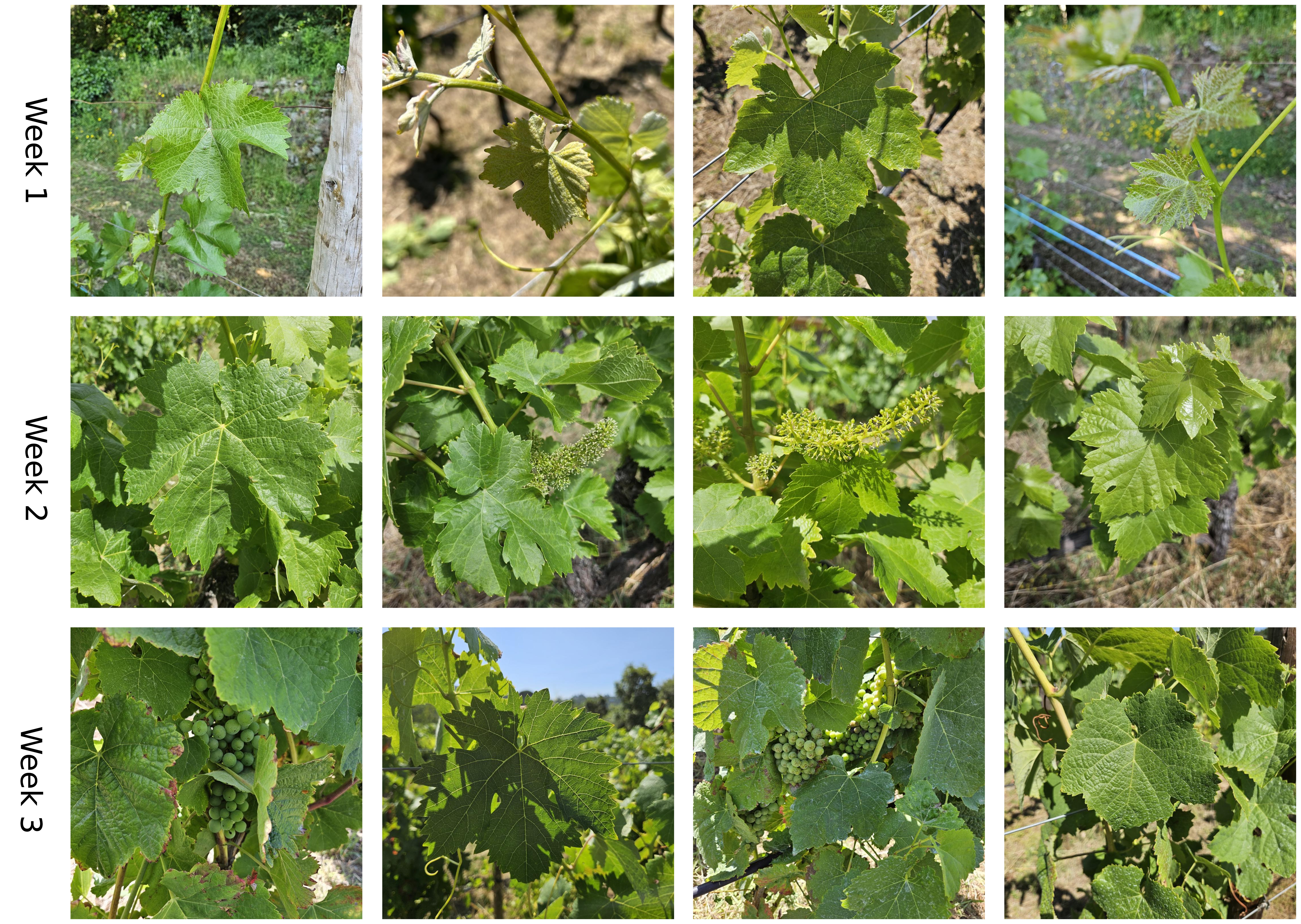}
	\caption{Examples of samples for each week in which data was acquired in Dataset 2. Each line represents a different week. Week 1 is composed of images of plants in leaf development and inflorescence emergence. Week 2 contains the majority of flowering images. Finally, in week 3 there are images of plants in the fruit development and berry ripening stages. \label{fig:dataset2_samples}}
\end{figure}

Dataset 2 (samples can be seen in \Cref{fig:dataset2_samples}) was designed to serve as a temporal extension of Dataset 1, capturing grapevine leaf images during the 2024 growing season. This dataset shares the same acquisition locations, grapevine varieties, and imaging parameters as Dataset 1, thereby maintaining consistency in the data collection protocol. It was specifically created to evaluate model performance across a complete new growing season and capture potential temporal variations in grapevine characteristics, enabling explicit evaluation of temporal robustness, a key challenge in grapevine variety classification since cultivar appearance changes markedly across the growing season \citep{chitwood_climate_2016}.

Images were acquired on seven dates in 2024, grouped into three periods, here called weeks. Week 1 includes 9, 23 and 24 May, Week 2 includes 12 and 20 June and Week 3 includes 16 and 17 July, covering various phenological stages, including leaf development, inflorescence emergence, flowering and early fruit development. Almost all the images (6,685 of 6,776) were acquired with an Apple iPhone 14 and a Samsung Galaxy S23, to ensure image quality consistency across the collection period. \Cref{fig:dataset2_samples} shows examples of images for the different weeks of acquisition, while the composition of the subsets can be seen in \Cref{fig:ds_composition}.

Dataset 2 was divided into training, validation and test subsets with 4,574, 1,060 and 1,142 samples, respectively. This dataset complements Dataset 1 by adding seasonal diversity to the training data, allowing for a more robust evaluation of the model’s ability to generalise across different phenological stages and environmental conditions. Dataset 2 is publicly available at \url{https://zenodo.org/records/15160944}.

\subsubsection{Unsupervised Grapevine Dataset 2023 (Dataset 3)}

\begin{figure}
	\includegraphics[width=\textwidth]{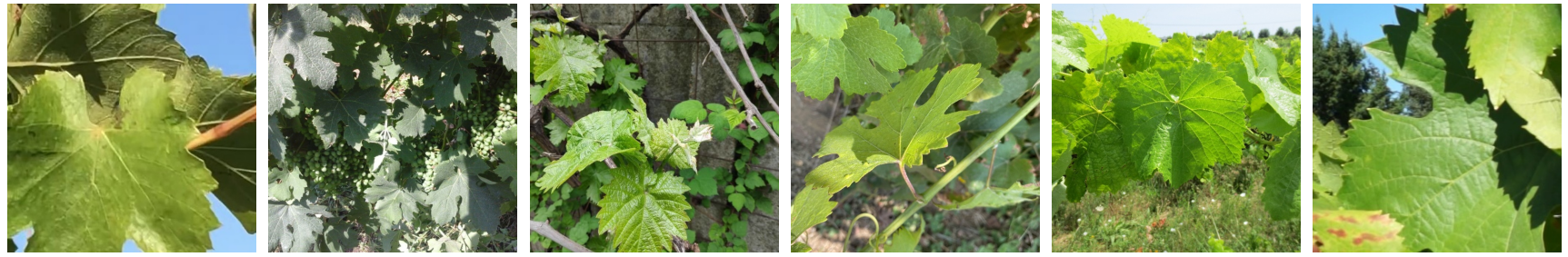}
	\caption{Dataset 3 examples. \label{fig:dataset3_samples}}
\end{figure}

Dataset 3 (see \Cref{fig:dataset3_samples}) consists of unlabelled grapevine leaf images collected from various sources \citep{hughes_open_2016, koklu_cnn-svm_2022, Vlah2021, lu_hybrid_2022, alessandrini_esca-dataset_2021, santos_grape_2020, moreira_gvxmi_2022, van_horn_benchmarking_2021}, including the training subset of Dataset 1 (images only, without labels; the validation and test subsets of Dataset 1 and all of Dataset 2 were excluded), and was used to pre-train models in a SSL context. Its distribution can be seen in \Cref{fig:ds_distribution} (left panel). Since the images were obtained from diverse datasets, a standardised square slicing technique was applied, allowing each original image to yield multiple fragments when feasible. Within each source, a consistent slicing pattern was maintained, while different patterns were applied across sources to manage redundancy, allowing up to 10\% overlap between slices to avoid generating visually redundant patches that could bias the reconstruction task. Square slicing increases the number of training samples while preserving local venation structures, which MAEs rely on during masked reconstruction. Augmented images were excluded, and only original samples were retained. For datasets containing video frames, frames were selected at 25-frame intervals to ensure diversity. In total, 33,671 images were included in this version of Dataset 3, with \gc{\Cref{fig:dataset3_samples}} showing sample images. It is important to highlight that no quality control was applied to the images. Regarding image duplication, a ResNet-50 model pre-trained on IN1K was used to extract visual features from images gathered through web crawling and from iNaturalist. To prevent duplication, cosine similarity was computed between feature vectors, and any images with a similarity score of 1 were identified as duplicates and removed from the dataset. This setup reflects the nature of unlabelled datasets that are often collected from multiple sources in practical applications.

\begin{figure}
	\centering
	\definecolor{dsBlue}{HTML}{2A78D6}\definecolor{dsOrange}{HTML}{EB6834}\definecolor{dsAqua}{HTML}{1BAF7A}%
	\definecolor{dsYellow}{HTML}{EDA100}\definecolor{dsMagenta}{HTML}{E87BA4}\definecolor{dsGreen}{HTML}{008300}%
	\definecolor{dsViolet}{HTML}{6250D6}\definecolor{dsRed}{HTML}{E34948}\definecolor{dsGrey}{HTML}{8D8C86}%
	\definecolor{dsLightGrey}{HTML}{C9C8C1}%
	\resizebox{\textwidth}{!}{%
	\begin{tikzpicture}
		\def\Rout{1.35}\def\Rin{0.85}
		\begin{scope}[shift={(0,0)}]
		\fill[dsBlue] (90.00:\Rout) arc[start angle=90.00, end angle=4.46, radius=\Rout] -- (4.46:\Rin) arc[start angle=4.46, end angle=90.00, radius=\Rin] -- cycle;
		\fill[dsOrange] (4.46:\Rout) arc[start angle=4.46, end angle=-80.42, radius=\Rout] -- (-80.42:\Rin) arc[start angle=-80.42, end angle=4.46, radius=\Rin] -- cycle;
		\fill[dsAqua] (-80.42:\Rout) arc[start angle=-80.42, end angle=-132.34, radius=\Rout] -- (-132.34:\Rin) arc[start angle=-132.34, end angle=-80.42, radius=\Rin] -- cycle;
		\fill[dsYellow] (-132.34:\Rout) arc[start angle=-132.34, end angle=-176.04, radius=\Rout] -- (-176.04:\Rin) arc[start angle=-176.04, end angle=-132.34, radius=\Rin] -- cycle;
		\fill[dsMagenta] (-176.04:\Rout) arc[start angle=-176.04, end angle=-213.44, radius=\Rout] -- (-213.44:\Rin) arc[start angle=-213.44, end angle=-176.04, radius=\Rin] -- cycle;
		\fill[dsLightGrey] (-213.44:\Rout) arc[start angle=-213.44, end angle=-270.00, radius=\Rout] -- (-270.00:\Rin) arc[start angle=-270.00, end angle=-213.44, radius=\Rin] -- cycle;
		\draw[white, line width=0.9pt] (90.00:\Rin) -- (90.00:\Rout);
		\draw[white, line width=0.9pt] (4.46:\Rin) -- (4.46:\Rout);
		\draw[white, line width=0.9pt] (-80.42:\Rin) -- (-80.42:\Rout);
		\draw[white, line width=0.9pt] (-132.34:\Rin) -- (-132.34:\Rout);
		\draw[white, line width=0.9pt] (-176.04:\Rin) -- (-176.04:\Rout);
		\draw[white, line width=0.9pt] (-213.44:\Rin) -- (-213.44:\Rout);
		\end{scope}
		\node[align=center, font=\small] at (0,0) {Dataset 3\\33,671\\images};
		\node[anchor=west, inner sep=0pt] at (1.6,0) {\scriptsize\renewcommand{\arraystretch}{1.05}%
			\begin{tabular}{@{}c@{\hspace{4pt}}l@{\hspace{6pt}}r@{}}
			\textcolor{dsBlue}{\rule{5pt}{5pt}} & \citet{al-khazraji_image_2023} & 23.76\% \\
			\textcolor{dsOrange}{\rule{5pt}{5pt}} & Dataset 1, training subset (own images) & 23.58\% \\
			\textcolor{dsAqua}{\rule{5pt}{5pt}} & \citet{lu_hybrid_2022} & 14.42\% \\
			\textcolor{dsYellow}{\rule{5pt}{5pt}} & \citet{hughes_open_2016} & 12.14\% \\
			\textcolor{dsMagenta}{\rule{5pt}{5pt}} & \citet{alessandrini_esca-dataset_2021} & 10.39\% \\
			\textcolor{dsLightGrey}{\rule{5pt}{5pt}} & Smaller sources & 15.71\% \\
			 & \quad Other sources & 4.48\% \\
			 & \quad \citet{moreira_gvxmi_2022} & 3.27\% \\
			 & \quad \citet{Vlah2021} & 3.00\% \\
			 & \quad Web crawling & 1.65\% \\
			 & \quad \citet{koklu_cnn-svm_2022} & 1.48\% \\
			 & \quad \citet{santos_grape_2020} & 0.94\% \\
			 & \quad iNaturalist \citep{van_horn_benchmarking_2021} & 0.89\%
			\end{tabular}};
		\begin{scope}[shift={(9.4,0)}]
		\fill[dsGrey] (90.00:\Rout) arc[start angle=90.00, end angle=-132.12, radius=\Rout] -- (-132.12:\Rin) arc[start angle=-132.12, end angle=90.00, radius=\Rin] -- cycle;
		\fill[dsOrange] (-132.12:\Rout) arc[start angle=-132.12, end angle=-211.39, radius=\Rout] -- (-211.39:\Rin) arc[start angle=-211.39, end angle=-132.12, radius=\Rin] -- cycle;
		\fill[dsGreen] (-211.39:\Rout) arc[start angle=-211.39, end angle=-230.58, radius=\Rout] -- (-230.58:\Rin) arc[start angle=-230.58, end angle=-211.39, radius=\Rin] -- cycle;
		\fill[dsViolet] (-230.58:\Rout) arc[start angle=-230.58, end angle=-246.06, radius=\Rout] -- (-246.06:\Rin) arc[start angle=-246.06, end angle=-230.58, radius=\Rin] -- cycle;
		\fill[dsRed] (-246.06:\Rout) arc[start angle=-246.06, end angle=-258.19, radius=\Rout] -- (-258.19:\Rin) arc[start angle=-258.19, end angle=-246.06, radius=\Rin] -- cycle;
		\fill[dsLightGrey] (-258.19:\Rout) arc[start angle=-258.19, end angle=-269.96, radius=\Rout] -- (-269.96:\Rin) arc[start angle=-269.96, end angle=-258.19, radius=\Rin] -- cycle;
		\draw[white, line width=0.9pt] (90.00:\Rin) -- (90.00:\Rout);
		\draw[white, line width=0.9pt] (-132.12:\Rin) -- (-132.12:\Rout);
		\draw[white, line width=0.9pt] (-211.39:\Rin) -- (-211.39:\Rout);
		\draw[white, line width=0.9pt] (-230.58:\Rin) -- (-230.58:\Rout);
		\draw[white, line width=0.9pt] (-246.06:\Rin) -- (-246.06:\Rout);
		\draw[white, line width=0.9pt] (-258.19:\Rin) -- (-258.19:\Rout);
		\end{scope}
		\node[align=center, font=\small] at (9.4,0) {Dataset 3+\\54,571\\images};
		\node[anchor=west, inner sep=0pt] at (11.0,0) {\scriptsize\renewcommand{\arraystretch}{1.05}%
			\begin{tabular}{@{}c@{\hspace{4pt}}l@{\hspace{6pt}}r@{}}
			\textcolor{dsGrey}{\rule{5pt}{5pt}} & Dataset 3 (left) & 61.70\% \\
			\textcolor{dsOrange}{\rule{5pt}{5pt}} & Self-acquired (own images) & 22.02\% \\
			\textcolor{dsGreen}{\rule{5pt}{5pt}} & \citet{sozzi_wgrapeunipd-dl_2022} & 5.33\% \\
			\textcolor{dsViolet}{\rule{5pt}{5pt}} & \citet{velez_precision_2023} & 4.30\% \\
			\textcolor{dsRed}{\rule{5pt}{5pt}} & \citet{morros_ai4agriculture_2021} & 3.37\% \\
			\textcolor{dsLightGrey}{\rule{5pt}{5pt}} & Smaller sources & 3.27\% \\
			 & \quad \citet{pinheiro_grapevine_2023} & 1.54\% \\
			 & \quad \citet{bertoglio_vineyard_2023} & 0.82\% \\
			 & \quad \citet{aguiar_grape_2021} & 0.71\% \\
			 & \quad \citet{wensheng_table_2022} & 0.20\%
			\end{tabular}};
	\end{tikzpicture}}
	\caption{Composition of Dataset 3 (left) and Dataset 3+ (right) by source, as a share of the images of each dataset. The images acquired by the authors are shown in orange, and the sources with smaller shares are grouped in grey.}
	\label{fig:ds_distribution}
\end{figure}

Although Dataset 3 aligns with the context of grapevine variety classification, it is not fully curated. Images exhibit variability in distance (greater than 40 cm or closer than 20 cm), and some images include non-leaf content, controlled-environment images, and images of leaves in suboptimal conditions. However, it is more contextually relevant to grapevine variety classification than natural scene datasets or multi-plant datasets. Dataset 3+ extends Dataset 3 by incorporating additional self-acquired images from vineyards near Vila Real, Portugal, that are different from the vineyards where Datasets 1 and 2 were acquired (so none of these images shows plants in the labelled datasets), as well as images from external sources \citep{sozzi_wgrapeunipd-dl_2022, bertoglio_vineyard_2023, velez_precision_2023, morros_ai4agriculture_2021, aguiar_grape_2021, pinheiro_grapevine_2023, wensheng_table_2022}, and includes 54,571 images in total.  Its distribution can be seen in \Cref{fig:ds_distribution} (right panel). The extended dataset, referred to as Dataset 3+, introduces more diverse content, including images of grape bunches, bare plants and plants photographed from a greater distance.

\subsection{Models Training}

\begin{figure}
	\includegraphics[width=\textwidth]{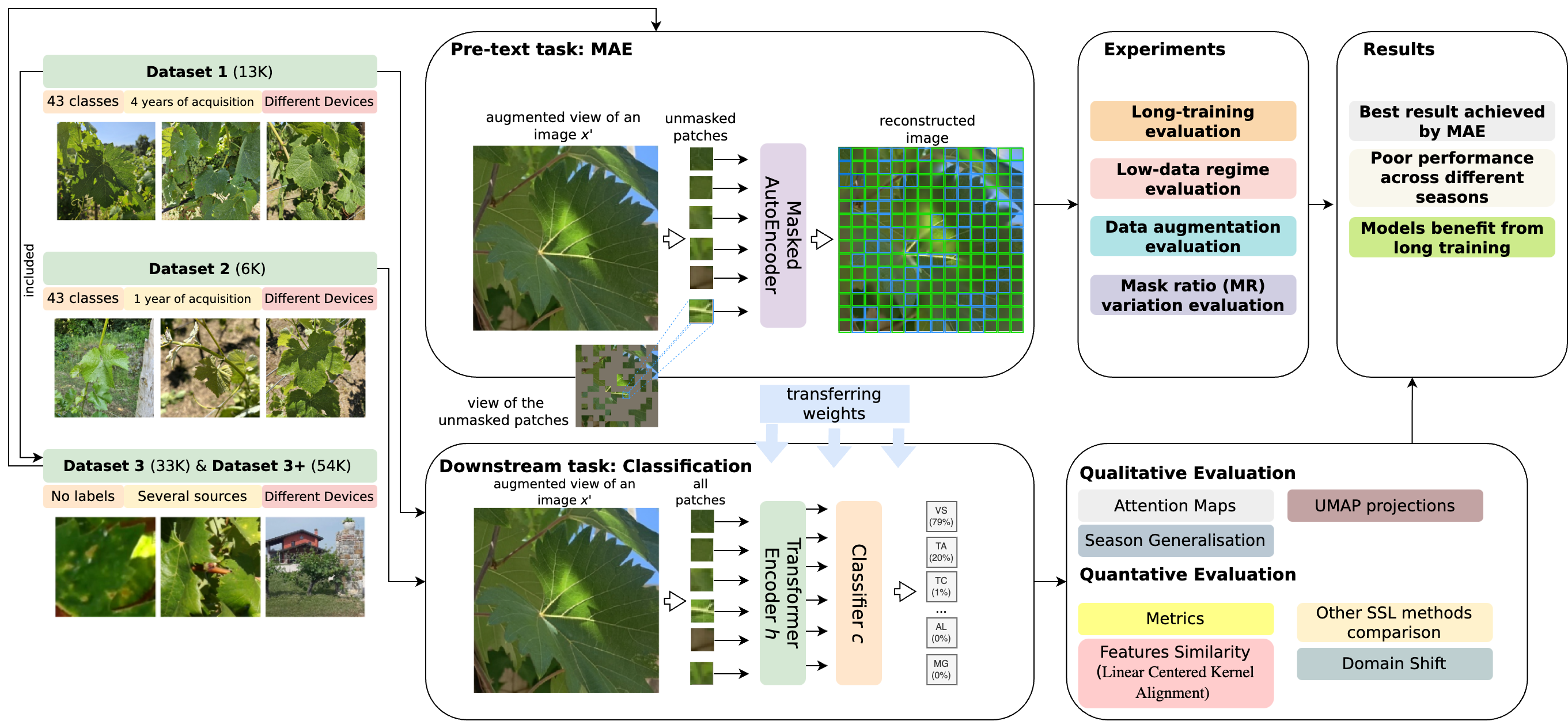}
	\caption{The study employs a two-phase training approach, with pre-text and downstream tasks. In the pre-text phase, an MAE reconstructs masked images using an unlabelled dataset. The decoder is then discarded. In the downstream phase, a classifier is trained on the encoder's top with an annotated dataset for grape variety classification. Evaluation includes classification metrics, similarity between features and attention maps. Furthermore, ablation studies on training duration, data availability, augmentation and mask ratios were carried out. \label{fig:training_approach}}
\end{figure}

This section describes the training procedures applied to the models used for grapevine variety classification. The approach combines ViT architectures and MAEs to maximise the efficacy of self-supervised pre-training and fine-tuning for the specific task of grapevine classification. \gc{\Cref{fig:training_approach}} gives an overview of the proposed method.

The training protocol consists of two main phases, namely pre-training (pre-text task) and fine-tuning (downstream task). During the pre-training phase, MAE is employed in an SSL setup using an unlabelled dataset (Dataset 3 or Dataset 3+). The MAE learns to reconstruct partially masked images, encouraging the model to extract meaningful representations without relying on labels. In the fine-tuning phase, the encoder is paired with a classification head and is fully retrained on labelled datasets (Dataset 1 and Dataset 2) to classify grapevine varieties. MAE was chosen because masked image modelling has been reported to perform well on fine-grained computer vision tasks \citep{ma_disjoint_2022}. Furthermore, MAE is computationally efficient, since the encoder processes only the visible patches (see \gc{\Cref{sec:mae}}). Considering recent studies \citep{huang_theoretical_2025, park_what_2023, assel_joint_2025, balestriero2024learningreconstructionproducesuninformative}, MAE tends to learn diverse and position-aware attention patterns. Each patch mainly attends to its own region and relies heavily on high-frequency information. This local specialisation is well suited for grapevine variety classification, where fine-grained patch-level details are important. However, the nature of the reconstruction pre-training task can lead to dimensional collapse, and features that are optimal for reconstruction may not transfer well to classification or detection tasks \citep{park_what_2023, assel_joint_2025}.

This two-phase approach leverages the self-supervised capabilities of MAEs to reduce dependency on annotated data, aligning with challenges in precision agriculture where labelled data are often limited. The following subsections detail the architectures used as well as the specific training parameters and configurations.

In essence, the Vision Transformers \citep{dosovitskiy_image_2020} models are used to classify grapevine varieties. However, instead of using pre-trained weights from IN1K dataset, MAE \citep{he_masked_2021} is applied to pre-train the models. After pre-training, the decoder is discarded and replaced by a classifier, keeping the obtained weights, and then the model is completely retrained.

\subsubsection{Masked Autoencoders (MAE)}\label{sec:mae}

MAEs were introduced by \citet{he_masked_2021}. The main idea behind MAE is to hide parts of the images and train an encoder-decoder architecture to reconstruct them. The patch-based nature of Vision Transformers \citep{dosovitskiy_image_2020} are explored to remove patches before feeding the encoder. The encoder processes only the unmasked patches and before the decoder stage, a placeholder token, $e_{mask}$, which is learned during training, is inserted into the positions of the removed patches. The decoder uses the processed tokens and the $e_{mask}$ to reconstruct the original image, being the loss calculated only on the patches that were hidden. \gc{\Cref{fig:training_approach}} shows an overview of this architecture. For a given image $x$, divided into $n$ patches of $d$ pixel values each, let $T$ be the set of masked patch indices and $x'$ the visible patches. The encoder produces latent representations $z' = h(x')$, and the decoder reconstructs all patches, producing $w = u(z', e_{mask})$. The loss is the mean squared error between the original pixel values $x$ and the reconstruction $w$, computed only on the masked patches, as described in \gc{\Cref{eq:mae}}. Pixel values were normalised with the IN1K channel mean and standard deviation, and the per-patch normalisation option of \citet{he_masked_2021} was not used.

\begin{equation}\label{eq:mae}
	\mathcal{L} = \frac{1}{|T|}\sum_{i=1}^{n} \mathbbm{1}_{[i \in T]} \left [ \frac{1}{d} \sum_{j=1}^{d} (x_{i,j} - w_{i,j})^2 \right ]
\end{equation}

\gc{\Cref{fig:training_approach}} illustrates the overall training methodology, which follows a two-phase process. First, input images are divided into patches, and a random subset of these patches is masked. Positional encodings are then added to the remaining visible patches, which are passed through the encoder. In the second phase, the decoder reconstructs the original image using both the encoder's output and $e_{mask}$ representing the masked patches. Importantly, the loss is computed only on the masked regions. This approach helps prevent feature collapse and reduces reliance on complex data augmentations, making MAEs an effective pre-training strategy, particularly for datasets with limited labelled data.

This study employs three ViT architectures as encoders, namely ViT-T/16, ViT-S/16 and ViT-B/16 \citep{dosovitskiy_image_2020, touvron_training_2021}. They differ primarily in their complexity, characterised by the number of attention heads and the embedding dimensions. ViT-T/16 is the smallest model, while ViT-B/16 has the largest capacity, with a higher number of heads and a larger embedding size, which improves its ability to learn from complex data. \gc{\Cref{tab:arch_details}} provides a summary of the architectural details for each model. All the models in this study used a patch size of 16 pixels and, for simplicity, will be referred to as ViT-T, ViT-S and ViT-B in the remainder of the text.

\begin{table}[]
	\caption{Summary of Vision Transformer architecture characteristics. The table provides details on patch size, embedding dimension, attention heads, number of Transformer blocks, and parameter count for each architecture (ViT-T/16, ViT-S/16, and ViT-B/16).}
	\label{tab:arch_details}
	\resizebox{\textwidth}{!}{%
		\begin{tabular}{@{}cccccc@{}}
			\toprule
			\textbf{Architecture} & \textbf{Patch Size}&\textbf{Embedding Dimension} & \textbf{Heads} & \textbf{Blocks} & \textbf{Parameters (M)} \\ \midrule
			ViT-T \citep{touvron_training_2021}& 16&192 & 3 & 12 & 5.80 \\
			ViT-S \citep{touvron_training_2021} & 16&384 & 6 & 12 & 22.20 \\
			ViT-B \citep{dosovitskiy_image_2020} & 16 & 768 & 12 & 12 & 86.00 \\ \bottomrule
		\end{tabular}%
	}
\end{table}

\section{Experimental Setup}
\label{sec:exp}
The main objective of this study is to assess the capability of MAEs to replace the commonly used IN1K transfer learning and fine-tuning approach in grapevine variety classification. Following the methodology proposed by \gc{\citet{he_masked_2021}}, models pre-trained with MAE were evaluated using a fine-tuning protocol, as the representations obtained through MAE serve as powerful non-linear features. After pre-training, the decoder was discarded and replaced with a fully connected ($C$) feedforward layer for classification purposes. A global average pooling layer was applied following the last block to generate the final representation for each input.

\subsection{Baselines}

The baseline models used in this study include ResNet-50 \citep{resnet}, Swin Transformer (Tiny and Base) \citep{liu_swin_2021}, ConvNextV2 (Tiny and Base) \citep{woo_convnext_2023}, DenseNet201 \citep{DBLP:journals/corr/HuangLW16a}, Xception \citep{DBLP:journals/corr/Chollet16a}, DEIT3 (Small and Base) \citep{touvron_deit_2022}, EfficientNetV2 (Small and Medium) \citep{DBLP:journals/corr/abs-2104-00298}, and ViT-S and ViT-B with supervised IN1K weights. These models were chosen based on their extensive use in prior studies \citep{de_nart_vine_2024, carneiro_grapevine_2022, magalhaes_toward_2023} or their good performance in general tasks.  All models were fine-tuned using a consistent procedure. For each architecture, the original classifier was removed and replaced with a fully connected layer sized to match the output features of the backbone, which was initialised with IN1K-pretrained weights. Training was performed over 100 epochs in two phases. During the first 10 epochs, the backbone remained frozen and only the newly added classifier was trained. After this warm-up period, the backbone was progressively unfrozen in three steps, with 30 percent of the layers unfrozen at epoch 10, 60 percent at epoch 30 and the entire backbone at epoch 60. Once fully unfrozen, all parameters were trained jointly until epoch 100. Throughout the entire process, the classifier used a learning rate ten times higher than the backbone, and a cosine decay schedule was applied over all epochs. This procedure provides a stable and reproducible fine-tuning strategy across different model architectures. The batch size was maintained at 32, and all hyperparameters were consistent with those used in the downstream task training (see \gc{\Cref{sec:dt_description}}). The supervised IN1K ViT-S and ViT-B were instead fine-tuned with the same protocol as the MAE pre-trained models (see \gc{\Cref{sec:dt_description}}).

Since the CNN, Swin and DeiT3 baselines and the MAE models were fine-tuned with different recipes (progressive unfreezing with a batch size of 32 against layer-wise learning-rate decay with SimCLR-style augmentation, CutMix and MixUp), the comparison between them must be considered carefully. The controlled comparison in this study is between ViT-B models that share the architecture and the fine-tuning recipe and differ only in the pre-training, namely the public MAE checkpoint pre-trained on IN1K by \citet{he_masked_2021} (referred to in the text as \emph{MAE-IN1K}, to distinguish it from the supervised IN1K baselines), the same checkpoint further pre-trained with MAE on Dataset 3 (\emph{MAE-IN1K $\rightarrow$ Dataset 3}), and MAE pre-training from random initialisation on Dataset 3 or Dataset 3+. In the tables and figures, where all ViT-B models listed under MAE were pre-trained with MAE, these two configurations are labelled \emph{IN1K} and \emph{IN1K $\rightarrow$ Dataset 3}.

\subsection{Pre-text task}

The implementation followed the setup provided in the original study by \citet{he_masked_2021}. In all experiments, an MAE decoder with 8 blocks and a representation size of 512 was stacked on the encoder. The AdamW optimiser was used, with a base learning rate of $1 \times 10^{-3}$ for experiments trained solely on Dataset 3/3+ and $1.5 \times 10^{-4}$ for models initialised with IN1K weights; following \citet{he_masked_2021}, the actual learning rate was obtained by scaling the base learning rate by the effective batch size divided by 256. The input size was set to $224 \times 224$ pixels, with batch sizes varying according to the encoder, namely 160 samples for ViT-T, 128 for ViT-S and 80 for ViT-B.

The mask ratio was kept at 0.60 in all main experiments (\gc{\Crefrange{tab:overall_results}{tab:domain_shift_weeks}}), except in the mask-ratio ablation. Weight decay was set to 0.05, with a 10-epoch warm-up period preceding a total training duration of 3,000 epochs. Pre-training checkpoints were saved at regular intervals; each was fine-tuned on Dataset 1, and the checkpoint reported in the main results was selected using the macro F1 score on the Dataset 1 validation subset. The test subsets were not used for model or checkpoint selection. The selected checkpoints were 2,500 epochs for ViT-T and ViT-S (Dataset 3), 2,000 epochs for ViT-B (Dataset 3) and 2,500 epochs for ViT-B (Dataset 3+). Other hyperparameters followed IN1K default settings \citep{he_masked_2021}. Training was conducted on a system equipped with an Intel Core i7-13700KF CPU, 48 GB of RAM, and an NVIDIA GeForce RTX 3060 GPU (12 GB). Random cropping was used as the sole data augmentation technique in all experiments, except for the data augmentation ablation. Dataset 3 and Dataset 3+ were used for the pre-text training phase.

For comparative purposes, SimSiam \citep{chen_exploring_2020}, Barlow Twins \citep{zbontar_barlow_2021} and DINO \citep{caron_emerging_2021} were also used to generate initial weights for ViT-S. Together, these methods cover a wide spectrum of self-supervised paradigms, which allows us to investigate how architectural and objective-function differences influence downstream performance.

DINO represents a contrastive approach over samples, where the model is encouraged to separate different images and align multiple views of the same one. Barlow Twins follows a different perspective and performs contrast over features, pushing the network to produce highly correlated embeddings of augmented views while discouraging redundancy across feature dimensions. SimSiam offers yet another view of contrastive learning by removing negative pairs entirely and relying on a stop-gradient mechanism to avoid collapse, making it simpler but also more sensitive to architectural and optimisation choices. Together, these contrasting philosophies help clarify how different pre-training dynamics shape the quality of representations for fine-grained grapevine variety classification.

The hyperparameters for these models matched those defined for IN1K, with an input size of $224 \times 224$ pixels, projector with $d = 2048$ and a batch size of 64. Local views were not used for DINO pre-training. The pre-text training task for these models was conducted over 1,000 epochs using Dataset 3. Remaining hyperparameters were kept at their defaults. ViT-S was chosen due to hardware constraints, as all experiments were conducted on a setup with only 12 GB of VRAM, which limits the size of models that could be trained. In this way, a preliminary comparison between the methods was made with the available computational resources.

\subsection{Downstream Task}\label{sec:dt_description}

In the fine-tuning phase, Datasets 1 and 2 were used to train the encoder specifically for the task of grapevine variety classification. Batch sizes remained consistent with those used during the pre-training phase, while the learning rate was set to $1 \times 10^{-3}$ with a decay factor of 0.65 applied to each layer. Training was performed over a total of 100 epochs, with the initial 10 epochs designated as a warm-up period.

Data augmentation was applied as per the SimCLR protocol \citep{chen_simple_2020}, which included random cropping, colour jitter, greyscale transformation and Gaussian blur. Additionally, CutMix and MixUp augmentations ($\alpha = 0.80$) were incorporated to increase sample diversity in the training set. Cross-Entropy Loss was employed as the objective function for fine-tuning.

\subsection{Evaluation}\label{sec:eval}

The models were evaluated across a range of scenarios, with the F1 Score used as the primary performance metric, complemented by Precision and Recall. For each class $c$, precision, recall and F1 score are computed one-versus-rest as in \gc{\Cref{eq:precision,eq:recall,eq:f1_score}}, where $TP_c$, $FP_c$ and $FN_c$ denote the true positives, false positives and false negatives for class $c$. All reported scores are macro averages over the $C=43$ classes. Macro precision and macro recall are the unweighted means of the per-class values, and the macro F1 score is the unweighted mean of the per-class F1 scores (\gc{\Cref{eq:macro}}), not the harmonic mean of macro precision and macro recall. Because each class has the same weight, macro recall is equal to balanced accuracy. Per-class recall, used in \gc{\Cref{tab:bad_classes}} to identify the most difficult classes, is the proportion of test images of class $c$ that are correctly classified.

\begin{equation}\label{eq:precision}
	Precision_c = \frac{TP_c}{TP_c + FP_c}
\end{equation}

\begin{equation}\label{eq:recall}
	Recall_c = \frac{TP_c}{TP_c + FN_c}
\end{equation}

\begin{equation}\label{eq:f1_score}
	F1_c = \frac{2 \cdot Precision_c \cdot Recall_c}{Precision_c + Recall_c}
\end{equation}

\begin{equation}\label{eq:macro}
	F1_{macro} = \frac{1}{C}\sum_{c=1}^{C} F1_c
\end{equation}

Since the mask ratio is a central hyperparameter of MAE, experiments were conducted with various mask ratios to assess its impact. Additionally, the initialisation of models with IN1K weights was tested to analyse its effect on convergence time and overall performance. For this purpose, models were pre-trained with MAE-IN1K weights as initial values for the MAE-based pre-training phase, and the resulting weights were used as starting points for the downstream fine-tuning task.

To further understand the model's behaviour, tests were conducted on long-duration pre-training, performance in low-data regimes, and the effects of applying stronger data augmentation during pre-training as proposed by \citet{chen_simple_2020}. Long pre-training was evaluated for all models, while low-data regime performance was assessed for the model that achieved the best overall results in the downstream task. For data augmentation analysis, ViT-B was pre-trained over 500 epochs with varying augmentation strategies to determine its impact on representation quality.

In addition to standard evaluation metrics, activation maps and pair-wise similarity between representations were analysed to compare the effects of IN1K initialisation and task-specific dataset pre-training on grapevine variety classification. Linear Centered Kernel Alignment (LCKA), as suggested by \citet{kornblith_similarity_2019}, was used to calculate similarity on 300 random images from the Dataset 1 test set, providing insights into the representational differences across models.

A domain shift analysis was also conducted to quantify the distributional differences between the datasets. Seasonal changes in the appearance of grapevine leaves can produce a substantial domain drift. Thus, domain shift analysis is useful to evaluate generalisation. The objective of this analysis was to assess how the representations learned during pre-training generalise to data collected under different conditions, particularly between the Dataset 1 and Dataset 2. This evaluation provides a complementary perspective to the model performance metrics, offering insights into the stability and adaptability of the learned feature spaces for different seasons.

To estimate the degree of domain shift, two complementary approaches were employed. The first method was based on the Maximum Mean Discrepancy (MMD), a kernel-based statistical measure that quantifies the distance between two probability distributions in a Reproducing Kernel Hilbert Space (RKHS), as described by \citet{DBLP:journals/corr/abs-0805-2368}. The Radial Basis Function (RBF) kernel was used. The unbiased estimator of $\mathrm{MMD}^2$ was computed, and its square root is reported. The bandwidth $\sigma$ was set once per encoder by the median heuristic (median pairwise distance between all features of that encoder; $\sigma = 21.1$ for Dataset 3+ and $\sigma = 2.9$ for MAE-IN1K) and was used for every comparison of that encoder. MMD values are therefore comparable across weeks, classes and reference comparisons of the same encoder. Significance was assessed with a permutation test with 200 permutations, so the smallest attainable $p$-value is 0.005. Features were the globally average-pooled outputs of the last block of the ViT-B encoders pre-trained with MAE on Dataset 3+ and of the MAE-IN1K encoder, without fine-tuning, computed on centre-cropped $224 \times 224$ images. Only the validation subsets were used (1,508 images from Dataset 1 and 1,060 from Dataset 2).

The second method followed a classifier-based discrimination approach. In this case, a binary logistic regression classifier was trained to distinguish between samples originating from the Dataset 1 and Dataset 2 using the same feature representations. The separability between the datasets was measured using the Area Under the Receiver Operating Characteristic Curve (AUC). The classifier was an L2-regularised logistic regression on standardised features. \gc{The AUC was computed with stratified five-fold cross-validation, so each image was scored by a classifier that did not see it during training. This held-out evaluation is necessary because the features have 768 dimensions, while each comparison has at most a few thousand images, and each per-class comparison has fewer than 100. With so many dimensions and so few images, a logistic regression can separate almost any two sets of images that it was trained on, even when the two sets come from the same distribution. For example, we split the Dataset 1 validation subset into two random halves, which have no shift between them. When the classifier was trained and evaluated on the same images, these two halves reached AUC values of 0.9526 and 0.9426 for the two encoders, and every per-class comparison reached 1.0000. These values measure how well the classifier memorises its training images and not the shift between the datasets. With held-out evaluation, the same two halves reached AUC values close to 0.5 (\Cref{tab:domain_shift_weeks}), as expected when there is no shift.} Both MMD and AUC are used as descriptive measures of distribution shift. The domain shift analysis was performed both globally and on a per-class basis, allowing the identification of specific grapevine varieties that are more sensitive to temporal or environmental variations. To support the interpretation of both measures, two reference comparisons were added. The first one uses two random halves of the Dataset 1 validation subset, for which no shift is expected, and the second one compares different seasons within Dataset 1 (2021 and 2022 against 2023), which are all present in the training data.

Finally, aiming to measure dimensional collapse within the features generated by MAE, recent studies \citep{zhuo_towards_2023, garrido_rankme_2023} were followed, and the effective rank \citep{roy_effective_2007} was applied to assess how the intrinsic dimensionality of the representations evolved during training across the different pre-training methods used in this study. The rank was computed from the activations of the final multi-head attention layer after global average pooling, using the validation sets of Dataset 1 and Dataset 2 as input. In addition, the features were projected using UMAP \citep{mcinnes_umap_2020} to visualise the feature spaces produced by MAE and the other self-supervised learning methods.

\section{Results}
\label{sec:results}
This section presents the results of the experiments conducted to evaluate the effectiveness of MAEs for grapevine variety classification. Detailed comparisons with baseline models, analyses of the impact of various training configurations, and results on dataset-specific performance are provided.

\subsection{Overall Performance and Baseline Comparison}

The performance of ViT models pre-trained with MAEs is compared to baseline models fine-tuned with IN1K weights. \gc{\Cref{tab:overall_results}} presents the precision, recall and F1 Scores for each model configuration. Among the supervised baseline models, DEIT3 (Base) achieved the highest F1 Score of 0.6817, followed closely by ConvNextV2 (Tiny) with an F1 Score of 0.6702. ViT-B pre-trained with MAE on Dataset 3+ reached an F1 Score of 0.7776, 0.0959 above DeiT3-B. However, these results must be considered carefully, since the supervised baselines and the MAE models were fine-tuned with different recipes (\gc{\Cref{sec:exp}}). Using the same architecture and fine-tuning recipe, the Dataset 3+ model outperformed MAE-IN1K (0.7494) by 0.0282. Since several test classes contain fewer than 20 images, small differences should be interpreted with caution. Precision, Recall and F1-Score per class for the ViT-B model pre-trained with MAE using Dataset 3+ and for MAE-IN1K, which achieved the best performance, are shown in \gc{\Cref{tab:app_complete_results}} (\gc{\Cref{app:perclass}}).

\begin{table}
	\centering
	\caption{Overall results of the experiments on Dataset 1. Macro-averaged precision, recall and F1 Score for each model configuration, comparing IN1K-based baselines with ViT models pre-trained using MAE on domain-specific data. Supervised baselines (except ViT-S and ViT-B) used the progressive-unfreezing recipe; all ViT models used the MAE fine-tuning recipe. In the MAE block, IN1K denotes the public MAE checkpoint of \citet{he_masked_2021}. Single run per configuration.}
	\label{tab:overall_results}
	\begin{tabular}{@{}cccc@{}}
		\toprule
		\textbf{Architecture} & \textbf{Precision} $\uparrow$ & \textbf{Recall} $\uparrow$ & \textbf{F1-Score} $\uparrow$ \\ \midrule
		\multicolumn{4}{c}{Baseline (Supervised IN1K)} \\ \midrule
		ResNet-50 & 0.5411 & 0.6404 & 0.5769 \\
		Xception & 0.5625 & 0.6255 & 0.5740 \\
		DenseNet201 & 0.6152 & 0.6656 & 0.6219 \\
		Swin Transformer  (Tiny) & 0.6471 & 0.7134 & 0.6659 \\
		ConvNextV2  (Tiny) & 0.6434 & 0.7229 & 0.6702 \\
		EfficientNetV2 (Small) & 0.4983 & 0.6093 & 0.5341 \\
		DEIT3    (Small) & 0.6370 & 0.7106 & 0.6639 \\
		ViT-S & 0.5702 & 0.6429 & 0.5909 \\
		Swin Transformer  (Base) & 0.6192 & 0.7357 & 0.6607 \\
		ConvNextV2  (Base) & 0.5998 & 0.6741 & 0.6214 \\
		EfficientNetV2  (Medium) & 0.5322 & 0.6269 & 0.5668 \\
		DEIT3    (Base) & 0.6521 & 0.7322 & 0.6817 \\
		ViT-B & 0.5475 & 0.6328 & 0.5658 \\ \midrule
		\multicolumn{4}{c}{MAE} \\ \midrule
		ViT-T (Dataset 3) & 0.4235 & 0.5174 & 0.4537 \\
		ViT-S (Dataset 3) & 0.6244 & 0.7117 & 0.6576 \\
		ViT-B (Dataset 3) & 0.6883 & 0.7628 & 0.7187 \\
		ViT-B (Dataset 3+) & \textbf{0.7597} & \textbf{0.8056} & \textbf{0.7776} \\
		ViT-B (IN1K) & 0.7297 & 0.7863 & 0.7494 \\
		ViT-B (IN1K $\rightarrow$ Dataset 3) & 0.7130 & 0.7695 & 0.7340 \\ \midrule
		\multicolumn{4}{c}{DINO} \\ \midrule
		ViT-S (Dataset 3) & 0.6240 & 0.6839 & 0.6446 \\ \midrule
		\multicolumn{4}{c}{BarlowTwins} \\ \midrule
		ViT-S (Dataset 3) & 0.4953 & 0.6007 & 0.5356 \\
		\bottomrule
	\end{tabular}
\end{table}

These results indicate that MAE pre-training on unlabelled, domain-specific images provides a competitive initialisation for this task. With ViT-B, pre-training on Dataset 3 (0.7187) came close to MAE-IN1K (0.7494) using about 38 times fewer images, and pre-training on Dataset 3+ exceeded it. The model hyperparameters had a large influence on the results, since, with the same pre-training data, ViT-T reached only 0.4537 and ViT-S 0.6576.

\begin{figure}
	\centering
	\includegraphics[width=0.6\textwidth]{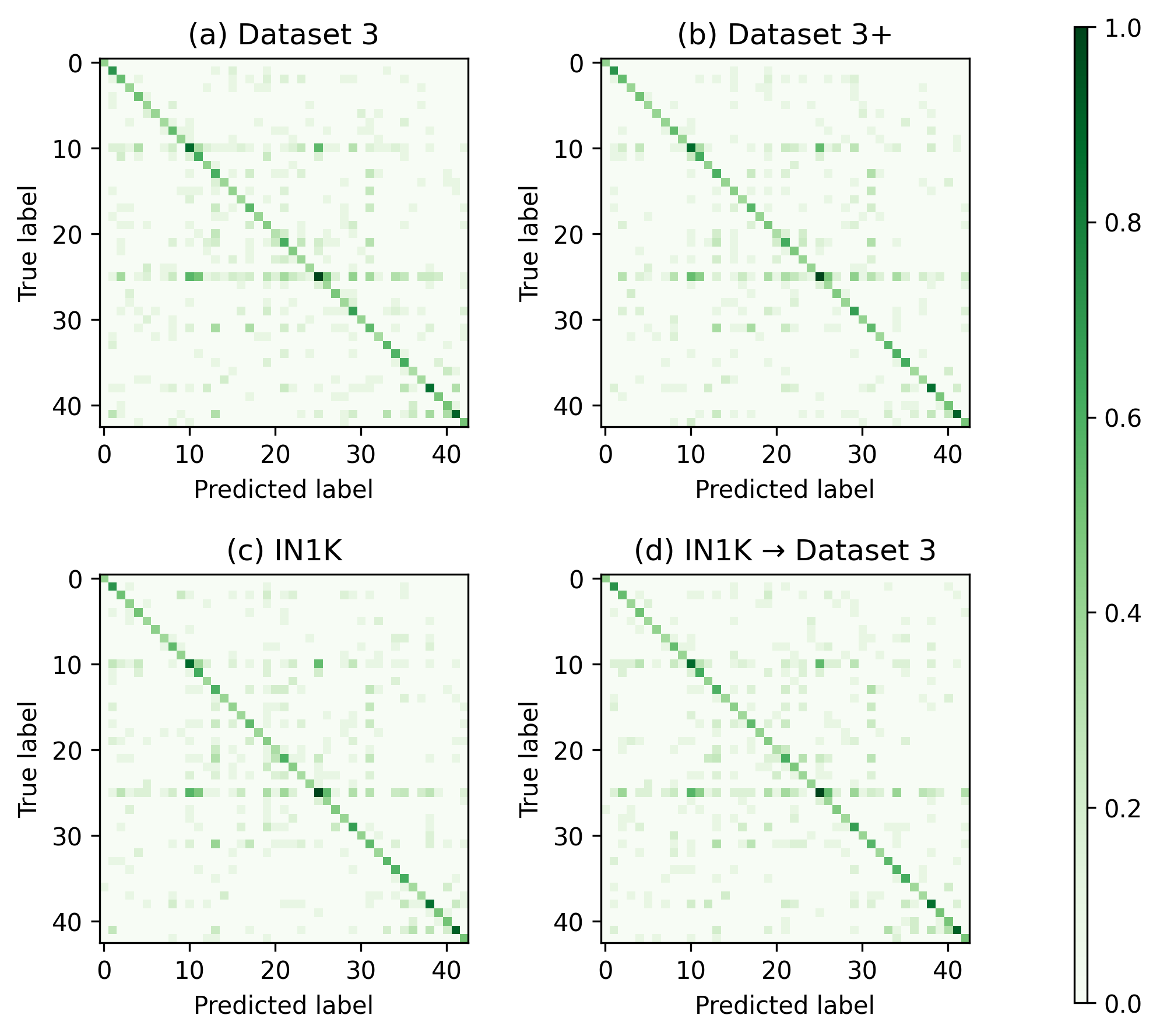}
	\caption{Confusion matrices for ViT-B models under different pre-training conditions, namely (a) Dataset 3, (b) Dataset 3+, (c) IN1K and (d) IN1K $\rightarrow$ Dataset 3. The class names are hidden for better visualisation and replaced by numerical indices ordered alphabetically. 0 represents the ‘Alicante’ class and 42 the ‘Viosinho’ class. Values were transformed using a logarithmic scale to enhance visualisation. \label{fig:cms}}
\end{figure}

To further understand class-specific performance across models, we analysed confusion matrices, as shown in \gc{\Cref{fig:cms}}. These matrices illustrate the distribution of true positives, false positives and misclassifications for each model configuration. The improvement of the Dataset 3+ model over MAE-IN1K is not uniform across classes, since it obtained a higher F1 score in 30 of the 43 classes and a lower one in 13 (\Cref{tab:app_complete_results}). For example, SM improved (F1 0.6258 against 0.5874), whereas EC became worse (0.5079 against 0.5758, with a recall of 0.5333 against 0.6333). For 28 of the 43 classes in each model (23 shared by both), recall exceeds precision, which is consistent with the much larger imbalance of the test subset compared with the training subset (\gc{\Cref{sec:dataset1}}), since the errors on the large test classes appear as false positives in the small ones.

To explore which specific classes posed the most difficulty for each model, \gc{\Cref{tab:bad_classes}} presents grapevine varieties with a recall below 0.7000 for the Dataset 3+ and MAE-IN1K models. These classes often represent varieties with close morphological similarities, leading to frequent misclassification by both models. Seven classes (EC, MC, MF, MR, MX, SM and TF) fall below this threshold for both models, indicating that they remain difficult regardless of the pre-training data. Domain-specific pre-training removed four classes from the list (CD, GM, GV and SB) but added three (CF, DT and FP); each of these three has fewer than 20 test images, so a single misclassified image changes its recall by more than 0.0500.

\begin{table}
	\centering
	\caption{Classes with a recall below 0.7000 on the Dataset 1 test subset for the ViT-B models pre-trained with MAE on Dataset 3+ and MAE-IN1K, both fine-tuned on Dataset 1. Per-class recall values are given in \Cref{tab:app_complete_results}. Classes below 0.7000 for both models are in bold. Class codes are defined in \Cref{fig:dataset1_samples}.}
	\label{tab:bad_classes}
	\begin{tabularx}{\textwidth}{@{}cX@{}}
		\toprule
		\textbf{Training Dataset} & \textbf{Classes} \\ \midrule
		Dataset 3+ & CF, DT, \textbf{EC}, FP, \textbf{MC}, \textbf{MF}, \textbf{MR}, \textbf{MX}, \textbf{SM}, \textbf{TF} \\ \addlinespace
		IN1K & CD, \textbf{EC}, GM, GV, \textbf{MC}, \textbf{MF}, \textbf{MR}, \textbf{MX}, \textbf{SM}, SB, \textbf{TF} \\ \bottomrule
	\end{tabularx}

\end{table}

Overall, the per-class results suggest that domain-specific pre-training helps most classes but does not resolve the confusions between the most similar varieties.

\subsection{Impact of Model Initialisation and Pre-training Configurations}

\begin{figure}
	\centering
	\includegraphics[width=0.85\textwidth]{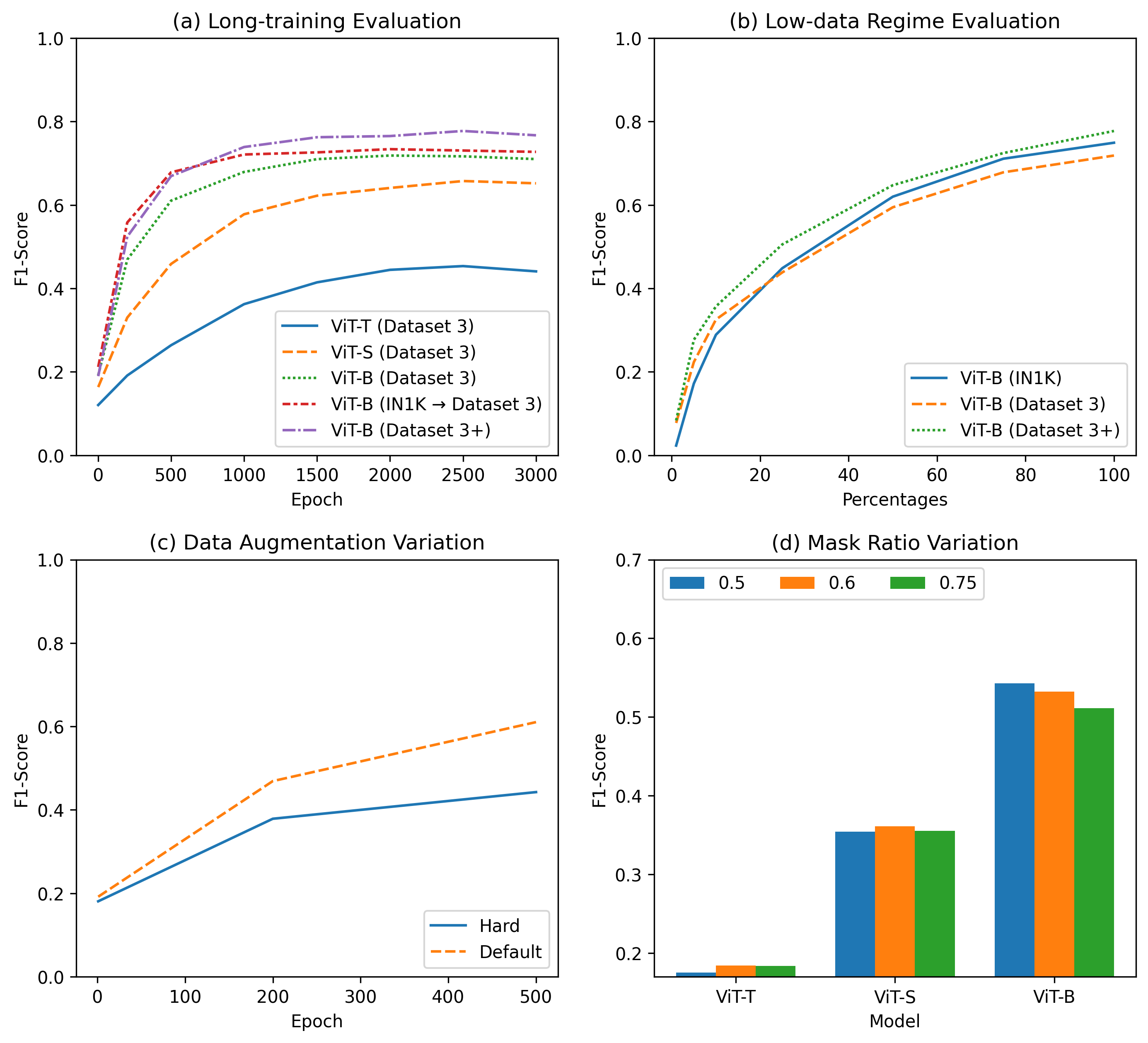}
	\caption{Evaluation of MAE pre-training with different configurations. (a) Convergence comparison for long-duration training (ViT-T, ViT-S and ViT-B), (b) Low-data regime performance (ViT-B), (c) Impact of strong data augmentation against random cropping (ViT-B, 500 epochs) and (d) Effect of mask ratio variations on model performance (ViT-T, ViT-S and ViT-B, 500 epochs). All values are Dataset 1 test F1 scores from single runs. \label{fig:ablations}}
\end{figure}

The initialisation of models with IN1K weights and different pre-training configurations were analysed to evaluate their impact on convergence speed, final model performance and data efficiency. \gc{\Cref{fig:ablations}(a)} presents a comparison of convergence patterns during the pre-training phase for ViT-T, ViT-S and ViT-B models pre-trained on Dataset 3, ViT-B pre-trained on Dataset 3+, and ViT-B initialised from MAE-IN1K and further pre-trained on Dataset 3. Each point is the Dataset 1 test F1 score obtained after fine-tuning the corresponding pre-training checkpoint; the curves are shown to illustrate convergence, while the checkpoints reported in \gc{\Cref{tab:overall_results}} were selected on the validation subset. Models initialised with MAE-IN1K weights exhibited faster early convergence within the first 500 epochs. In the controlled comparison, which uses the same pre-training data (Dataset 3), MAE-IN1K initialisation remained slightly ahead of random initialisation at the selected checkpoints (0.7340 versus 0.7187), but both stayed below the MAE-IN1K checkpoint fine-tuned directly (0.7494). The higher score of the Dataset 3+ model (0.7776) changes both the pre-training data and the initialisation at the same time, so it cannot be attributed to the initialisation. These results suggest that IN1K initialisation speeds up early training, and that the amount of in-domain pre-training data matters more than the starting point.

To further explore the impact of pre-training on model efficiency in low-data regimes, we tested the performance of ViT-B models on progressively smaller subsets of labelled data during fine-tuning. As shown in \gc{\Cref{fig:ablations}(b)}, models pre-trained on Dataset 3+ outperformed those initialised with MAE-IN1K weights at every labelled-data fraction tested, including scenarios where only 10\% of the labelled data was used for fine-tuning. Since each fraction was evaluated with a single run, the consistent direction of the difference is more reliable than the difference in each fraction. This result suggests that MAE pre-training on in-domain data is label efficient, which is particularly valuable in domains like viticulture, where labelled datasets are often costly and time-consuming to obtain.

The effect of data augmentation strategies was also examined to determine the optimal approach during pre-training. \gc{\Cref{fig:ablations}(c)} illustrates the performance impact of strong data augmentation (e.g., heavy random cropping and colour distortion) compared to simpler transformations such as random cropping alone. Strong augmentation led to a reduction in F1 Score by 27.4\% in ViT-B, suggesting that complex augmentations may introduce noise that is detrimental to feature learning in this context. This finding aligns with the observation of \citet{he_masked_2021} on IN1K that MAE works well with minimal augmentation.

Lastly, the effect of varying the mask ratio was evaluated to assess its influence on model performance. The results in \gc{\Cref{fig:ablations}(d)} indicate that the mask ratio had a small impact, with the best values at 0.6 for ViT-T and ViT-S models and 0.5 for ViT-B. The sweep covered ratios between 0.50 and 0.75 after 500 pre-training epochs, and the mask ratio had only a small effect on the F1 score. The main experiments used 0.60 for all architectures.

\subsection{Qualitative and Feature Similarity Analysis of Representations}

To further understand how pre-training configurations affect the learned representations, qualitative analyses were conducted by examining attention maps and representation similarity across different configurations. These analyses provide insights into the model's ability to capture relevant features specific to grapevine varieties, particularly in distinguishing subtle morphological differences.

\begin{figure}
	\centering
	\includegraphics[width=0.8\textwidth]{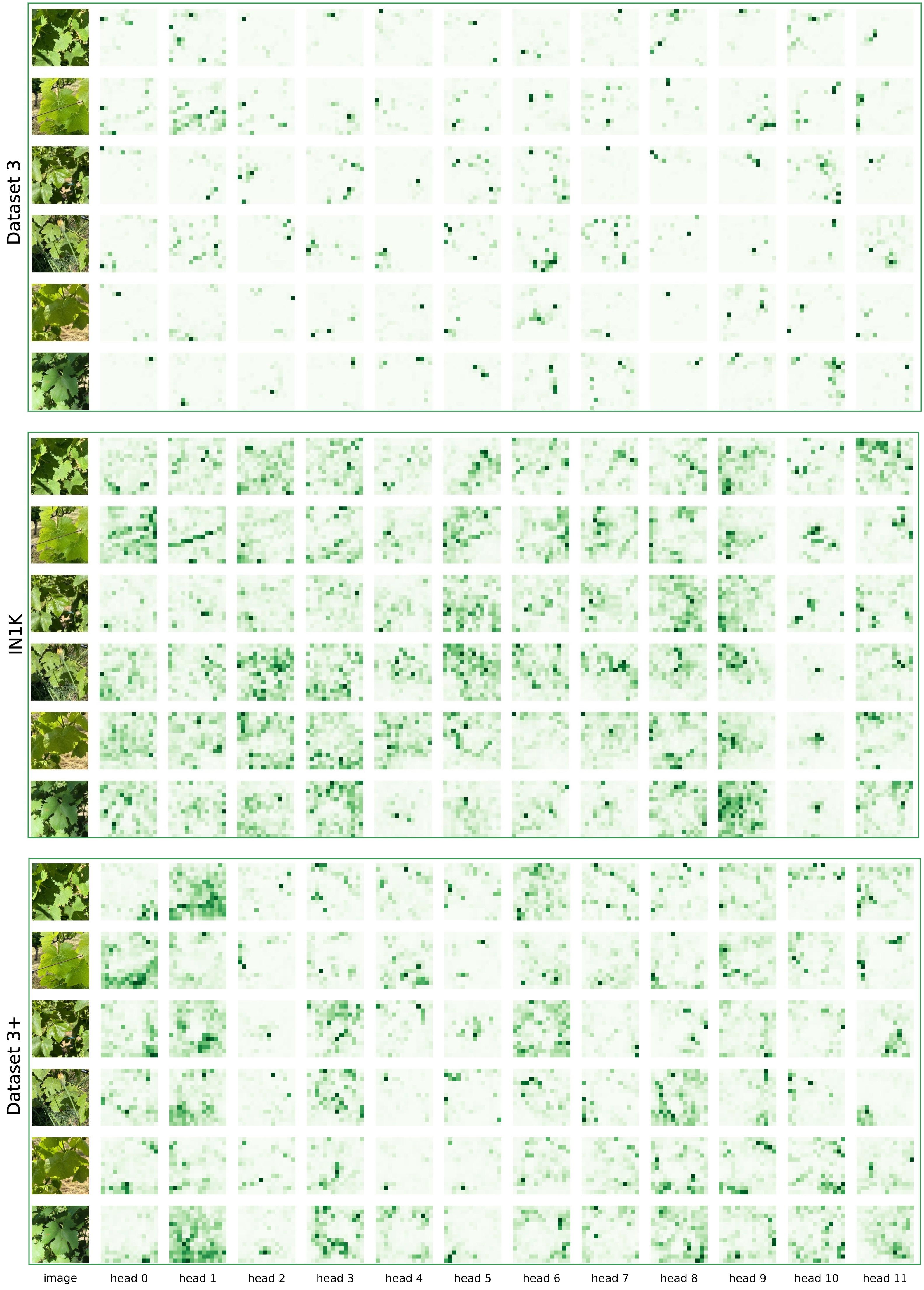}
	\caption{Attention maps for ViT-B models pre-trained on Dataset 3, IN1K and Dataset 3+, and fine-tuned on Dataset 1, for six randomly chosen test images. The attention maps were obtained using the output of each head belonging to the last multi-head attention block. \label{fig:attn}}
\end{figure}

\gc{\Cref{fig:attn}} displays attention maps for six random samples, comparing ViT-B models pre-trained on Dataset 3, MAE-IN1K and Dataset 3+. These maps are presented as qualitative illustrations. In these examples, the heads of the Dataset 3+ model tend to attend to leaf structures such as veins, edges and lobes, while some heads of the MAE-IN1K model attend to larger regions that include background.

\begin{figure}
	\centering
	\includegraphics[width=0.7\textwidth]{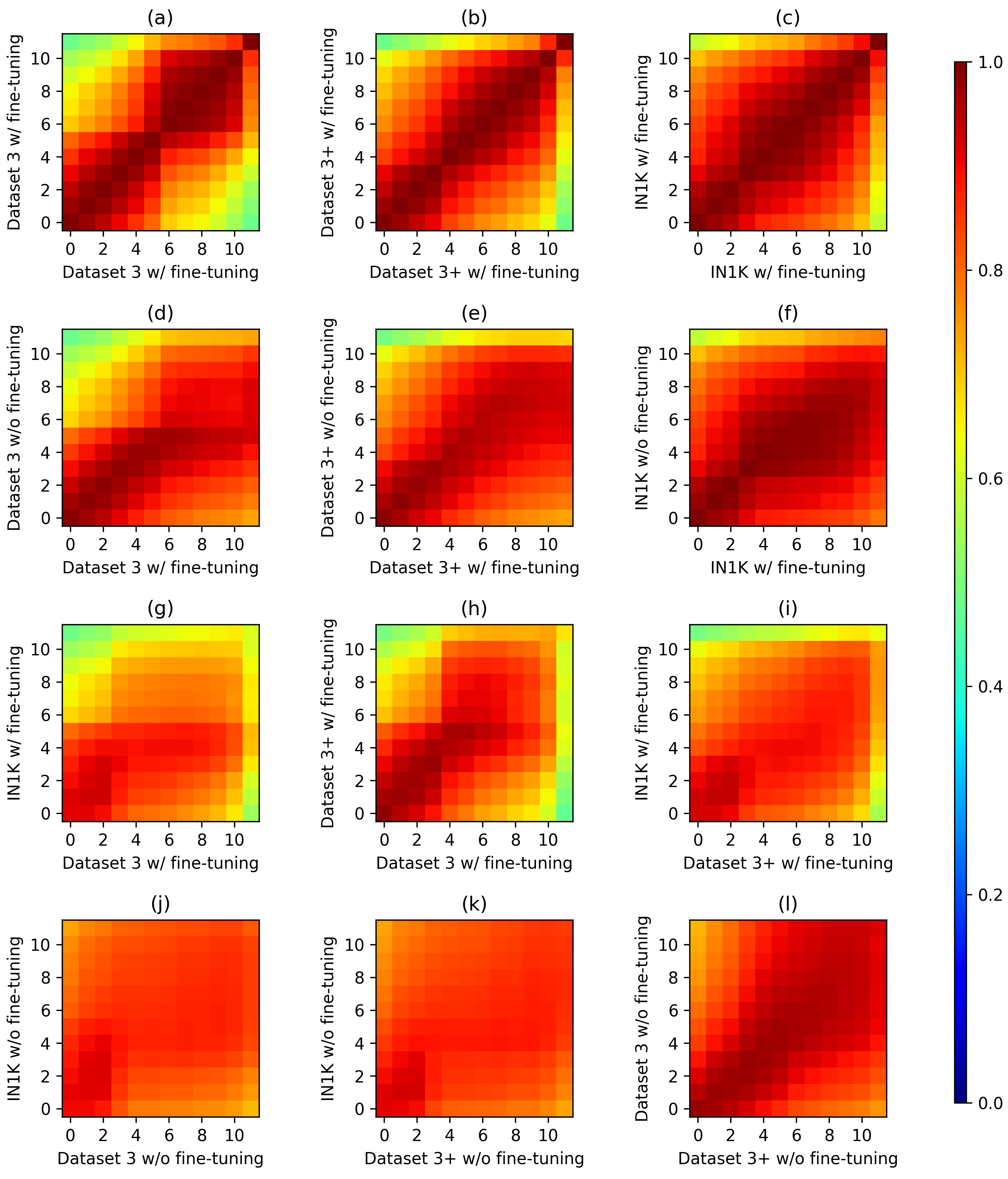}
	\centering
	\caption{Pairwise similarity for the experiments using Dataset 3, Dataset 3+ and IN1K in the pre-text task.  Similarity was obtained using LCKA \citep{kornblith_similarity_2019} for 300 random images from the test subset of Dataset 1, and the heatmap was composed using the average of the similarity of the representations of the output of each ViT-B multi-head attention block. The x and y axes represent the index of the blocks and their labels the datasets used for comparison and whether fine-tuning was performed using Dataset 1.  \label{fig:linear_cka}}
\end{figure}

In addition to attention maps, pairwise similarity between learned representations was analysed to evaluate the diversity of features captured in different layers. \gc{\Cref{fig:linear_cka}} shows similarity heatmaps for ViT-B models pre-trained on Dataset 3, MAE-IN1K and Dataset 3+, calculated on 300 random images from the test subset of Dataset 1. The models pre-trained on Dataset 3+ show a higher degree of dissimilarity across layers, especially in the higher layers, meaning that consecutive upper blocks transform the representation more than in the other models.

In comparison, the MAE-IN1K model's similarity heatmap shows a more uniform pattern across its upper blocks.

\subsection{Generalisation and Seasonal Data Performance}\label{sec:seasonal}

To assess the ability of MAE pre-trained models to generalise to new seasonal data, ViT-B models were evaluated on Dataset 2, which includes data collected in a different growing season. This test provides insights into the models' robustness when encountering temporal variations in grapevine morphology due to seasonal growth patterns. As shown in \gc{\Cref{tab:res_test_weekly}}, all models lost a large part of their performance on the new season. The F1 score of the Dataset 3+ model decreased from 0.7776 on the Dataset 1 test subset to 0.4677 on the Dataset 2 test subset (a relative drop of 39.9\%), and that of MAE-IN1K decreased from 0.7494 to 0.4747, a difference of 0.0070 in favour of MAE-IN1K. Therefore, domain-specific pre-training did not improve the generalisation to a new season.

\begin{table}
	\centering
	\caption{F1 Scores of ViT-B models pre-trained on various datasets (Dataset 3, Dataset 3+, IN1K, IN1K $\rightarrow$ Dataset 3), fine-tuned on Dataset 1 only, and evaluated on seasonal data (Dataset 2). Weeks 1 to 3 and All use all Dataset 2 images of the corresponding period; Test Subset uses only the Dataset 2 test subset. Week-by-week analysis shows reduced scores in the initial week, likely due to limited feature differentiation in early growth stages.}
	\label{tab:res_test_weekly}
	\begin{tabular}{@{}cccccc@{}}
		\toprule
		\textbf{Training Dataset} & \textbf{Week 1} $\uparrow$ & \textbf{Week 2} $\uparrow$ & \textbf{Week 3} $\uparrow$ & \textbf{All} $\uparrow$ & \textbf{Test Subset} $\uparrow$ \\ \midrule
		Dataset 3 & 0.3461 & 0.4595 & 0.4165 & 0.4104 & 0.4120 \\
		Dataset 3+ & \textbf{0.3909} & 0.5102 & 0.4984 & 0.4685 & 0.4677 \\
		IN1K & 0.3674 & \textbf{0.5144} & \textbf{0.5189} & \textbf{0.4698} & \textbf{0.4747} \\
		IN1K $\rightarrow$ Dataset 3 & 0.3496 & 0.5120 & 0.4678 & 0.4444 & 0.4425 \\ \bottomrule
	\end{tabular}
\end{table}

When analysing performance by acquisition week within Dataset 2, however, we observe a notable challenge in the first acquisition week, focused on leaf development stages. During this period, all models recorded F1 Scores below 0.4000, likely due to the high morphological similarity across grapevine varieties at early phenological stages, where distinguishing features are less pronounced, and to the fact that Dataset 1 contains few images from this stage. Early growth stages therefore require additional training data that cover them.

Further experiments were conducted by merging the training subsets of Datasets 1 and 2 for fine-tuning. As shown in \gc{\Cref{tab:retraining}}, this improved the F1 score on both test subsets for all models, with the ViT-B model pre-trained on Dataset 3+ achieving the highest scores (0.7956 on Dataset 1 and 0.8112 on Dataset 2). The gain on Dataset 1 (0.0180 to 0.0357) shows the benefit of adding images from a new season, although part of it may also be related to the larger number of labelled images (58\% more).

\begin{table}
	\centering
	\caption{Results after merging the training subsets of Dataset 1 and Dataset 2 into a single dataset and fine-tuning the ViT-B models pre-trained on Dataset 3, Dataset 3+, IN1K and IN1K $\rightarrow$ Dataset 3. The results are the F1-Score for the test subsets of Dataset 1 and Dataset 2 and, in brackets, the change compared to fine-tuning the models using Dataset 1 only (\gc{\Cref{tab:overall_results}} for Dataset 1 and the Test Subset column of \gc{\Cref{tab:res_test_weekly}} for Dataset 2).}
	\label{tab:retraining}
	\begin{tabular}{@{}ccc@{}}
		\toprule
		\textbf{Training Dataset} & \textbf{Dataset 1} $\uparrow$ & \textbf{Dataset 2} $\uparrow$ \\ \midrule
		Dataset 3 & 0.7425 (+0.0238) & 0.7769 (\textbf{+0.3649}) \\
		Dataset 3+ & \textbf{0.7956} (+0.0180) & \textbf{0.8112} (+0.3435) \\
		IN1K & 0.7851 (\textbf{+0.0357}) & 0.7794 (+0.3047) \\
		IN1K $\rightarrow$ Dataset 3 & 0.7629 (+0.0289) & 0.7824 (+0.3399) \\ \bottomrule
	\end{tabular}
\end{table}

For applications in precision agriculture, where varieties need to be classified over different growing seasons, these results indicate that labelled data from the target season are more important than the choice of pre-training.

Complementing the seasonal generalisation experiments, a domain shift analysis was conducted to quantify the distributional differences between Datasets 1 and 2, namely between the samples of the 2024 season and those of the previous seasons. We employed both kernel-based and classifier-based metrics. MMD was computed with an RBF kernel and a binary classifier was used (see \gc{\Cref{sec:eval}}). The results are shown in \gc{\Cref{tab:domain_shift_weeks}} for the whole of Dataset 2, for each week and within Dataset 1, and in \gc{\Cref{tab:domain_shift}} (\gc{\Cref{app:perclass}}) per class, for the ViT-B encoders pre-trained with MAE on Dataset 3+ and MAE-IN1K, the models that achieved the best performance.

As expected, the two random halves of Dataset 1 could not be distinguished (AUC of 0.5290 and 0.5090, MMD close to zero). In contrast, Dataset 2 was almost perfectly separable from Dataset 1 in all the weeks and for both encoders (AUC between 0.9819 and 0.9990), which indicates a large shift between the 2024 season and the previous ones. \gc{Since Dataset 2 was acquired mostly with two smartphone models, part of this shift may come from the acquisition devices and not only from the season.} The AUC values of the two encoders were very similar, so this shift is a property of the data and not of the pre-training. The MMD decreased from Week 1 to Week 3 for both encoders (0.2606, 0.2265 and 0.1569 for Dataset 3+, and 0.3446, 0.2068 and 0.1455 for MAE-IN1K), which shows that the early-season images are the most distant from Dataset 1, in agreement with the lower F1 scores in Week 1 (\Cref{tab:res_test_weekly}). The seasons within Dataset 1 were also separable from each other (AUC between 0.9477 and 0.9999), which indicates that each growing season has its own characteristics. \gc{Therefore, the AUC shows that each acquisition campaign is distinct, while the MMD indicates which periods are the most distant.} Since these seasons are part of the training data, the models can learn this variability, while the 2024 season was never seen during training.

\begin{table}
	\centering
	\caption{Domain shift between Dataset 1 and Dataset 2 (per week and whole set) and within Dataset 1, for the ViT-B encoders pre-trained with MAE on Dataset 3+ and IN1K (without fine-tuning), computed on the validation subsets. AUC is the held-out AUC of a logistic regression domain classifier (five-fold cross-validation), where 0.5 means that the two sets cannot be distinguished, and MMD is the square root of the unbiased MMD$^2$, with one RBF bandwidth per encoder. All comparisons except the random halves have $p < 0.01$ (MMD permutation test), and $n$ is the number of images in each set.}
	\label{tab:domain_shift_weeks}
	\begin{tabular}{@{}lccccc@{}}
		\toprule
		\multirow{2}{*}{\textbf{Comparison}} & \multirow{2}{*}{$n$} & \multicolumn{2}{c}{\textbf{Dataset 3+}} & \multicolumn{2}{c}{\textbf{IN1K}} \\ \cmidrule(l){3-6}
		& & \textit{AUC} $\downarrow$ & \textit{MMD} $\downarrow$ & \textit{AUC} $\downarrow$ & \textit{MMD} $\downarrow$ \\ \midrule
		\multicolumn{6}{@{}l}{\textit{Dataset 1 against Dataset 2}} \\
		Week 1 & 1,508 / 382 & 0.9990 & 0.2606 & 0.9970 & 0.3446 \\
		Week 2 & 1,508 / 334 & 0.9957 & 0.2265 & 0.9855 & 0.2068 \\
		Week 3 & 1,508 / 344 & 0.9898 & 0.1569 & 0.9917 & 0.1455 \\
		Whole Dataset 2 & 1,508 / 1,060 & 0.9858 & 0.1653 & 0.9819 & 0.1842 \\ \midrule
		\multicolumn{6}{@{}l}{\textit{Within Dataset 1}} \\
		Random halves & 754 / 754 & 0.5290 & 0.0000 & 0.5090 & 0.0072 \\
		Season 2022 against 2023 & 290 / 947 & 0.9999 & 0.2421 & 0.9781 & 0.1526 \\
		Season 2021 against 2023 & 190 / 947 & 0.9841 & 0.2110 & 0.9477 & 0.2175 \\ \bottomrule
	\end{tabular}
\end{table}

Per-class results in \gc{\Cref{tab:domain_shift}} (\gc{\Cref{app:perclass}}) show a significant shift ($p < 0.05$) for 41 of the 43 classes with the Dataset 3+ encoder and for 40 with MAE-IN1K; the exceptions were Cabernet Sauvignon and Chardonnay, and also Viosinho for MAE-IN1K. The held-out AUC ranged from 0.7593 to 1.0000 (mean 0.9291) for Dataset 3+ and from 0.6725 to 1.0000 (mean 0.8943) for MAE-IN1K, and the MMD from 0.1041 to 0.3189 and from 0.1045 to 0.3830, respectively. The severity of the shift was not uniform across varieties, and the two encoders ranked the classes similarly (Spearman correlation of 0.7391 for MMD and 0.8454 for AUC). Cabernet Sauvignon, Viosinho and Chardonnay had among the smallest MMD values with both encoders, while Cabernet Franc, Pinot Blanc, Fernão Pires, Códega do Larinho, Malvasia Corada and Malvasia Rei were among the ten largest for both. These differences may be related to the phenological stages covered or to changes in leaf morphology between seasons \citep{chitwood_climate_2016}. The varieties with a recall below 0.7000 for both models on Dataset 1 (Esgana C\~ao, Malvasia Corada, Malvasia Fina, Malvasia Rei, Moscatel Galego Roxo, Samarrinho and Tinta Francisca; \Cref{tab:bad_classes}) do not share a particular level of shift. With the Dataset 3+ encoder, their MMD ranges from among the lowest (Esgana C\~ao) to the highest (Malvasia Rei), and Malvasia Corada, Malvasia Rei and Moscatel Galego Roxo are among the twelve largest values for both encoders. Since these errors were measured on the Dataset 1 test subset, they are not caused by the shift to 2024, and are more likely related to morphological similarity between varieties and intra-class variability.

Overall, these results show that neither pre-training approach generalises well to the 2024 season, and that both encoders separate the 2024 images from the previous seasons to the same extent, with the largest shift in the early-season images.

\subsection{Comparison with other SSL Methods}

Different SSL methods lead to varying performances depending on the architecture used. Grapevine variety classification is a fine-grained task, requiring features capable of capturing subtle differences between images. In this study, ViT-S models were pre-trained using MAE, DINO and Barlow Twins, while SimSiam was also tested but failed to converge, exhibiting severe collapse\gc{, so it was not included in the comparison}.

\begin{figure}
	\centering

	\begin{subfigure}{0.30\textwidth}
		\centering
		\includegraphics[width=\linewidth]{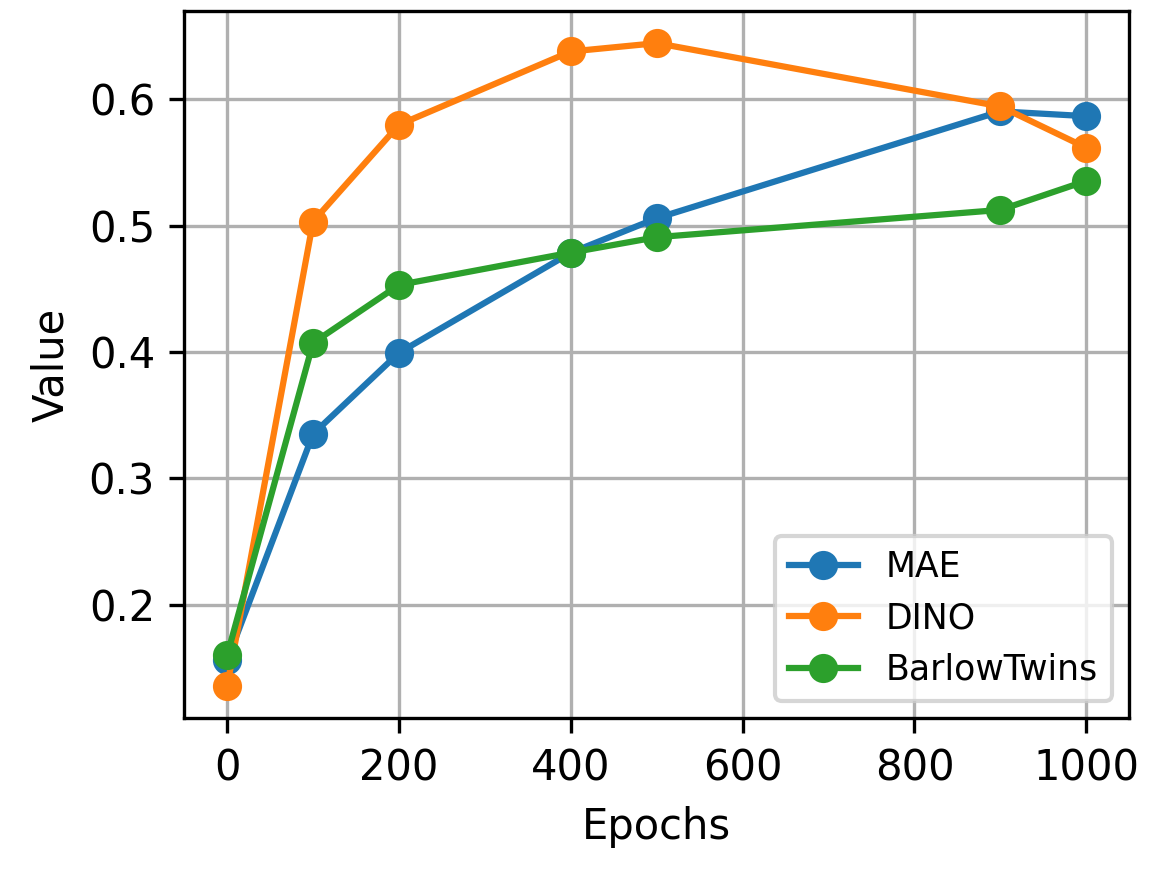}
		\caption{Fine-tuning results for MAE, DINO and Barlow Twins pre-trained on Dataset 3 over 1000 epochs.}
		\label{fig:sub1}
	\end{subfigure}
	\hfill
	\begin{subfigure}{0.30\textwidth}
		\centering
		\includegraphics[width=\linewidth]{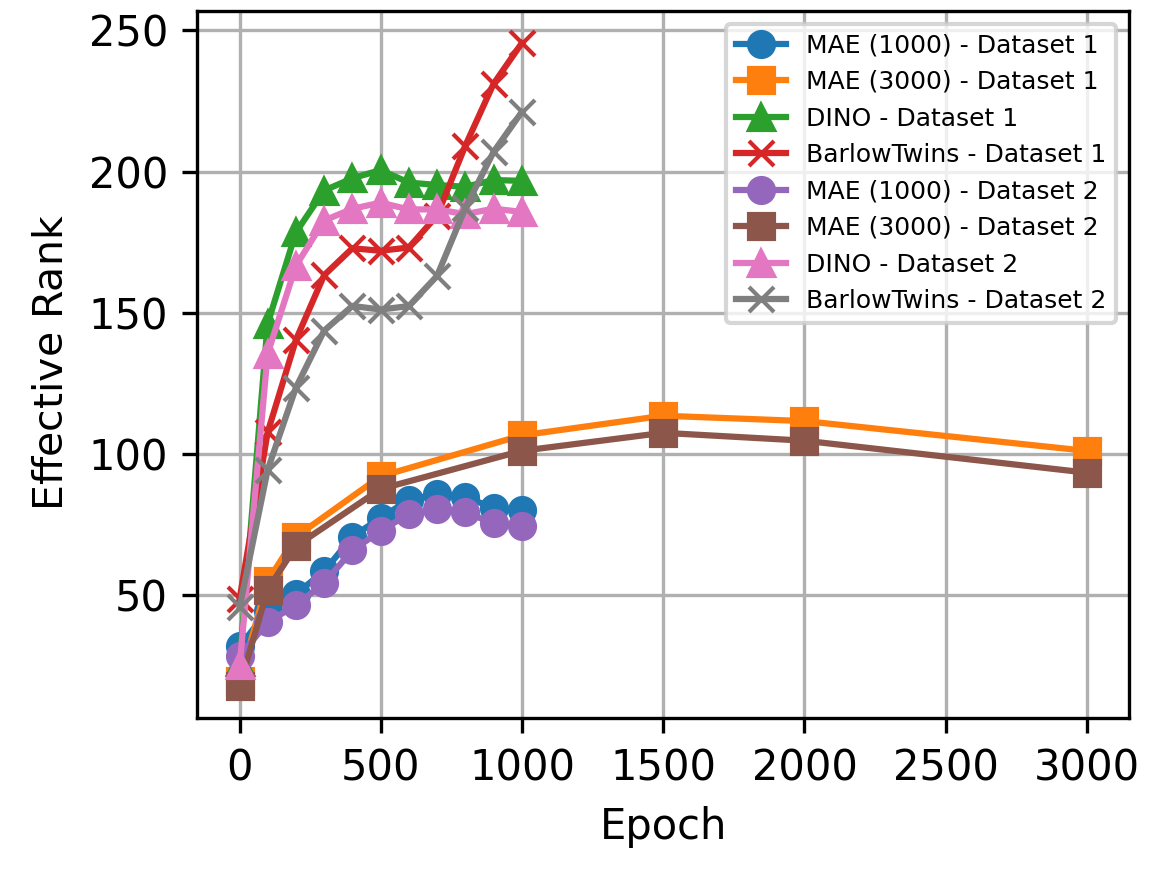}
		\caption{Effective rank for MAE, DINO and Barlow Twins computed on the validation subsets of Dataset 1 and Dataset 2.}
		\label{fig:sub2}
	\end{subfigure}
	\hfill
	\begin{subfigure}{0.30\textwidth}
		\centering
		\includegraphics[width=\linewidth]{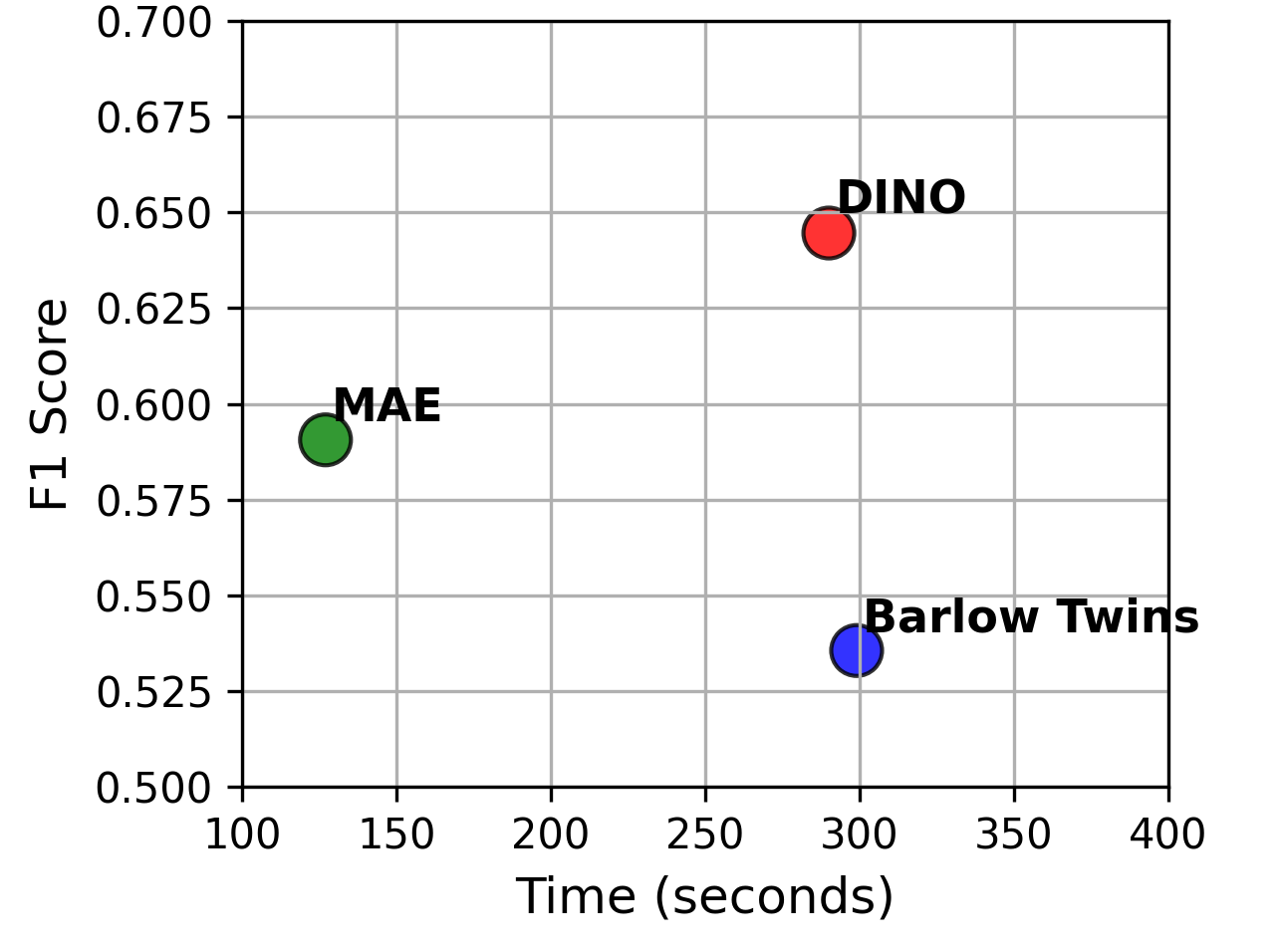}
		\caption{Pre-training time per epoch against F1 score for MAE, DINO and Barlow Twins.}
		\label{fig:sub3}
	\end{subfigure}

	\caption{Results for ViT-S pre-trained with MAE, DINO and Barlow Twins on Dataset 3.}
	\label{fig:ssl_methods_results}
\end{figure}

These SSL methods represent three distinct pre-training strategies, namely reconstruction-based (MAE), sample-contrastive (DINO) and feature-contrastive (Barlow Twins). \Cref{fig:sub1} summarises the fine-tuning performance over the epochs. Considering that in previous experiments MAE was trained for 3000 epochs, whereas DINO and Barlow Twins were trained for only 1000 epochs, MAE was retrained under the same conditions as the other methods, matching both the number of epochs and batch size.

\paragraph{Overall Performance.} The results show that DINO converges the fastest among the three methods. By 100 epochs, its F1 score had already reached 0.5034, surpassing both Barlow Twins at 0.4073 and MAE at 0.3352. DINO's performance gradually increased with further epochs, reaching 0.6446 at 500 epochs, where it began to plateau; this is the checkpoint reported in \gc{\Cref{tab:overall_results}}.

MAE started slower, with an F1 score of 0.3352 at 100 epochs, and improved steadily, reaching 0.5868 at 1000 epochs. At this matched budget of 1000 epochs, DINO therefore obtained the best result (0.6446 against 0.5868). MAE only matched DINO when its pre-training was extended to 3000 epochs (0.6576 against 0.6446, \gc{\Cref{tab:overall_results}}).
Barlow Twins exhibits intermediate behaviour, converging faster than MAE but slower than DINO. Its performance plateaus around 400 epochs, and it obtained the lowest final F1 score of the three methods.

Overall, these results highlight a trade-off between convergence speed and the cost per epoch. Under our settings, DINO gave the best result at a matched number of epochs, while MAE needed a longer schedule to reach a similar level but was cheaper per epoch. Barlow Twins, although relatively stable, does not match MAE's peak performance or DINO's early convergence. These findings illustrate the different characteristics of self-supervised learning strategies in the context of fine-grained grapevine variety classification, as shown in \Cref{fig:ssl_methods_results}.

However, it is important to consider the computational costs associated with each method. One epoch of DINO with the same hyperparameters as MAE requires more than twice the GPU time (MAE = 127 s; DINO = 290 s), as shown in \Cref{fig:sub3}. This highlights a critical trade-off between computational efficiency and model performance that must be considered when selecting an appropriate method for specific applications.

\paragraph{Effective Rank Analysis.} The concept of effective rank is used to quantify the effective dimensionality of a representation learned by a model. In essence, it measures how many principal directions in the feature space capture most of the variance, providing insight into how efficiently the model encodes information. \Cref{fig:sub2} shows that contrastive methods generally achieve a higher effective rank than MAE, and this behaviour is consistent across both Dataset 1 and Dataset 2. For DINO, the effective rank closely mirrors its performance, increasing steadily and then plateauing starting around epoch 500. Barlow Twins achieves the highest effective rank among all methods, which is expected since its redundancy reduction term in the loss function encourages information to be spread across more dimensions. MAE exhibits a similar pattern when trained for 1000 and 3000 epochs, although longer training allows the model to use more dimensions in its representations, resulting in a higher effective rank. The low effective rank of MAE indicates that its representations use only a small part of the available dimensions, a behaviour known as dimensional collapse \citep{zhuo_towards_2023, garrido_rankme_2023}. This is a disadvantage for a fine-grained domain such as grapevine variety classification, where the differences between classes are subtle and a large number of features is needed to separate them. However, the high effective rank of Barlow Twins did not result in a better F1 score, which shows that the diversity of the features alone is not enough for this task.

\begin{figure}
	\centering
	\includegraphics[width=1\textwidth]{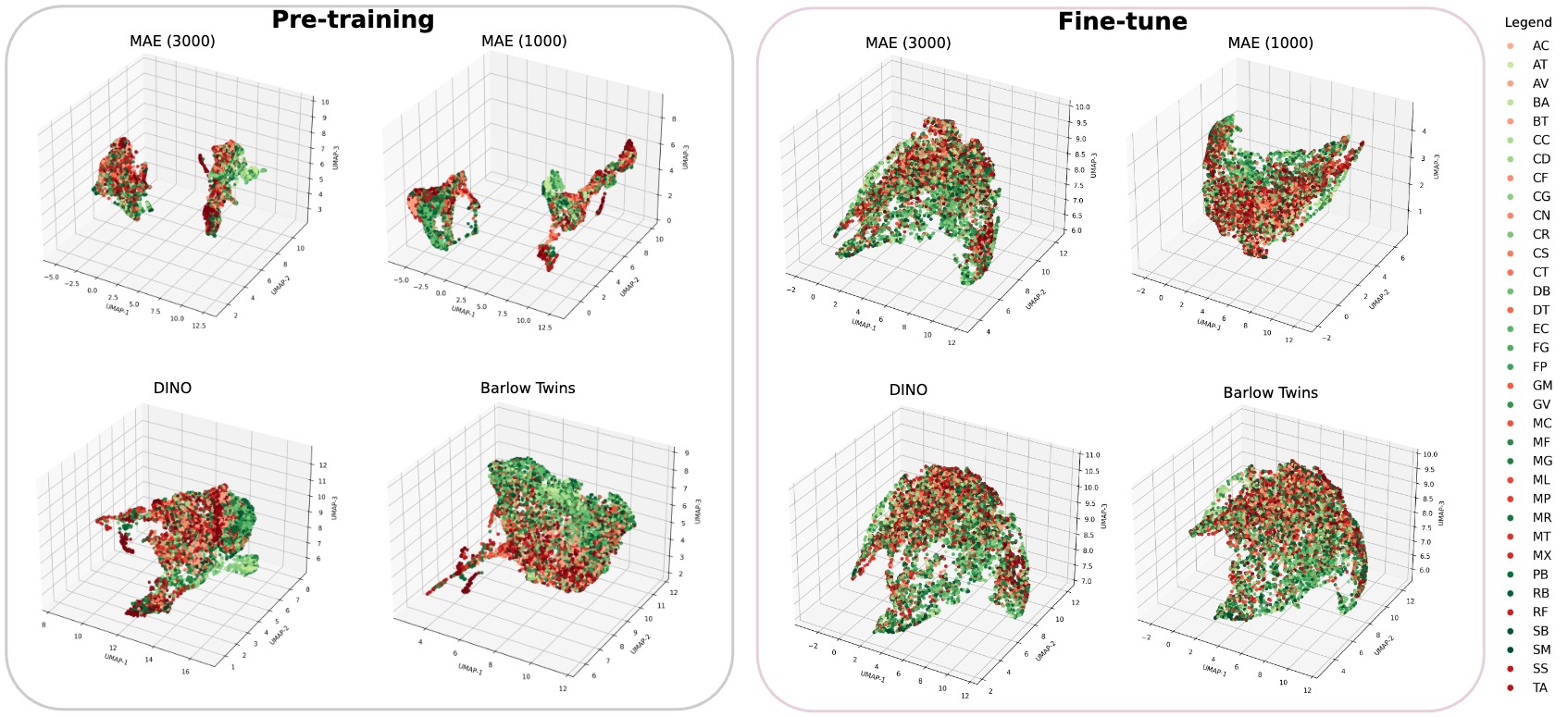}
	\caption{UMAP embeddings of Dataset 1’s validation set for ViT models pre-trained with MAE (1000 and 3000 epochs), DINO and Barlow Twins (left), and their corresponding fine-tuned versions (right). \label{fig:met_umap}}
\end{figure}

\paragraph{UMAP Visualisation.} \Cref{fig:met_umap} presents the 3D UMAP projections of the features extracted from the validation set of Dataset 1 for each pre-training method, both before and after fine-tuning. In the projections, the pre-trained MAE features appear more concentrated than those of the contrastive methods\gc{, and they form two separate groups. However, this difference cannot be used as a measure of dimensionality. UMAP keeps the local neighbourhoods of the points, but it does not keep the global distances or the variance of the features. Therefore, the size of the projected clouds does not show how many dimensions the features use, and the lower effective rank of MAE is supported by \Cref{fig:sub2} and not by the UMAP projections.} Clusters corresponding to the grapevine varieties can be observed for all methods, and after fine-tuning the projections of all methods look more similar.

\section{Discussion}
\label{sec:discussion}
This study investigates the potential of MAEs as a pre-training method for grapevine variety classification, comparing it to supervised transfer learning approaches using IN1K weights. The findings suggest that MAE pre-training on domain-specific datasets (Dataset 3 and Dataset 3+) yields a competitive and label-efficient initialisation for this task, but its benefit does not extend to a new season, where all initialisations obtained similar results. Here, we discuss these results and place them in the context of the existing literature.

\subsection{Benefits of Domain-Specific MAE Pre-training}

The best performance on the Dataset 1 test subset was achieved by the model pre-trained on Dataset 3+, 0.0282 above MAE-IN1K with the same fine-tuning recipe. On the 2024 season (Dataset 2), however, MAE-IN1K performed slightly better (0.4747 against 0.4677), so the benefit of domain-specific pre-training did not extend to the new season. Notably, Dataset 3, which is 38 times smaller than IN1K, demonstrated that a smaller, specialised pre-training dataset can achieve performance comparable to IN1K. Increasing the pre-training data from Dataset 3 to Dataset 3+ (62\% more images) increased the F1 score by 8.2\% and 13.5\% in relative terms on the test subsets of Dataset 1 and Dataset 2, respectively. Since the added images come from vineyards and sources that are not part of the labelled datasets, this gain is not explained by overlap with the evaluation data.

Another important point is that neither unlabelled dataset was curated, and Dataset 3+ contains a larger share of off-target content, such as fruit, sky or plant trunks without leaves. Nonetheless, the larger and less clean Dataset 3+ produced better results, which suggests that, at this scale, the quantity and diversity of in-domain images matter more than their cleanliness. In the similarity heat maps (\Cref{fig:linear_cka}), the Dataset 3+ model also shows lower similarity between neighbouring upper blocks.

This study aligns with others \citep{guldenring_self-supervised_2021, min_htet_strawberry_2023, yousra_self-supervised_2023} that report advantages of in-domain SSL pre-training over supervised IN1K pre-training in agricultural applications.

\subsection{Impact of using IN1K Initialisation}

Continuing MAE pre-training on Dataset 3 from the MAE-IN1K checkpoint gave a lower score (0.7340) than fine-tuning MAE-IN1K directly (0.7494), but a higher one than pre-training on Dataset 3 from random initialisation (0.7187). During the first 500 pre-training epochs, IN1K initialisation led to the best performance among all trained models. This indicates that IN1K initialisation accelerates convergence, but that further pre-training on a small in-domain dataset, with the hyperparameters used here, reduced part of the benefit of the IN1K checkpoint.

It is worth noting that the same hyperparameters were used for both random initialisation and IN1K initialisation as starting weights during pre-training. Adjusting the hyperparameters for IN1K initialisation could potentially improve performance. These findings contrast with those of \citet{guldenring_self-supervised_2021} in long training scenarios but are consistent when fewer than 1,000 epochs are considered. Their study demonstrated that using IN1K initialisation in ResNet-18 led to better performance on dense tasks when pre-trained for 600 epochs using SwAV \citep{caron_unsupervised_2021}.

\subsection{Ablation Experiments}

All pre-trained models benefited from long pre-training, though their downstream performance began to decline before reaching 3,000 epochs. These findings align with those of \citet{he_masked_2021}, who observed improvements in IN1K performance up to 1,600 epochs without overfitting when using ViT-L \citep{dosovitskiy_image_2020}. The results further suggest that the benefits of long pre-training are influenced by both the dataset size and the model's parameters. For Dataset 3, the best performance was achieved at 2,500 epochs for ViT-T and ViT-S, and at 2,000 epochs for ViT-B. In contrast, using Dataset 3+, ViT-B achieved optimal performance at 2,500 epochs, indicating that increasing the dataset size delays overfitting.  To the best of our knowledge, only \citet{hindel_inod_2023} have evaluated the effects of long pre-training in agricultural contexts, concluding that SSL pre-training benefits from more than 800 epochs. Other studies applying MAE in agriculture \citep{wang_classification_2024, liu_self-supervised_2022, kang_identification_2023, li_research_2024} have not examined pre-training beyond 500 epochs.

In a low-data regime, Dataset 3 outperformed MAE-IN1K when up to 10\% of annotated data was used in the downstream task, and Dataset 3+ outperformed all other models at every proportion of data tested. These findings are consistent with other SSL studies in agriculture \citep{guldenring_self-supervised_2021, david_dong_self-supervised_2023, hindel_inod_2023}, and suggest that pre-training with a vineyard-specific dataset is a good choice when annotated data is limited.

Regarding data augmentation and mask ratio, stronger augmentations led to worse results compared to random cropping, consistent with findings for IN1K \citep{he_masked_2021}. Mask ratio had a minor impact on performance. For ViT-T and ViT-S, the best results were obtained with a mask ratio of 0.60, while ViT-B performed best at 0.50. The main experiments used 0.60 for all architectures. These tests, conducted over 500 epochs for ratios between 0.50 and 0.75, may not generalise to longer training scenarios. Comparisons with IN1K \citep{he_masked_2021} show a best masking ratio of 0.60 for fine-tuning, though tested with ViT-L. Other studies applying MAE in agricultural contexts \citep{wang_classification_2024, liu_self-supervised_2022, kang_identification_2023, li_research_2024} have not explored different mask ratio. The results suggest that, within the range tested, the mask ratio has only a slight influence on performance.

\subsection{Seasonal Generalisation and the Role of Diverse Data}

A significant limitation of publicly available image datasets for grapevine variety classification \citep{koklu_cnn-svm_2022, al-khazraji_image_2023, santos_embrapa_2019, sozzi_wgrapeunipd-dl_2022, seng_computer_2018, Vlah2021} is their narrow scope, since they typically include only a few varieties and are collected over a single day or growing season \citep{carneiro_deep_2024}. Beyond discussing SSL methods, this study introduces, to our knowledge, the largest public benchmark dataset for grapevine variety identification in number of varieties, and presents a model capable of classifying 43 varieties with an F1 Score of 0.7956 when trained on the merged Datasets 1 and 2 (0.7776 when trained on Dataset 1 only).

The results obtained by evaluating models trained on Dataset 1 against Dataset 2 (see \gc{\Cref{tab:res_test_weekly}}) highlight the importance of seasonal diversity in training datasets. The F1 score of the Dataset 3+ model trained on Dataset 1 decreased from 0.7776 to 0.4677 when it was tested on the 2024 season, even with the same sites and protocol, and the other initialisations showed similar drops. This indicates that datasets acquired in one or few seasons may not represent all the variability found in the field. Merging Datasets 1 and 2 improved the test performance on Dataset 1 for all models, which shows the benefit of adding images from a new season. These findings emphasise the need for more varied image acquisition in terms of time, location and grapevine variety in order to develop classifiers suited for practical applications.

To the best of current knowledge, only \citet{de_nart_vine_2024} validated a deep learning model on an external dataset \citep{Vlah2021}, reporting accuracy values between 29.00\% and 35.00\% for different architectures. Although their setting differs from ours, both studies show that the performance decreases when models are applied to images acquired in different conditions.

Further analysis with Dataset 2 made it possible to identify the phenological stages that posed the greatest difficulty for the models (see \gc{\Cref{tab:res_test_weekly}}). The stages near leaf development and inflorescence emergence were consistently the most challenging for distinguishing varieties. This observation aligns with earlier work showing that morphological similarities among grapevine varieties are most pronounced during early phenological stages \citep{chitwood_latent_2016, macklin_intraspecific_2022}, which reduces the availability of distinctive visual features. The weakness observed in Week 1 reflects this issue clearly. At this stage, leaves show reduced morphological differentiation and appear more homogeneous across cultivars, which limits the discriminative cues available to the models. This effect appears across all pre-training strategies, including MAE pre-training on the domain-specific dataset. The domain shift analysis is consistent with this interpretation, since Week 1 forms the most distinct distribution relative to Dataset 1 (largest MMD for both encoders).

The domain shift analysis showed that the 2024 images are clearly separated from the previous seasons in the feature spaces of both the Dataset 3+ and the MAE-IN1K encoders, with the largest shift in the early-season images, which may be related to inter-annual changes in leaf morphology \citep{chitwood_climate_2016}\gc{ or to the acquisition devices, since Dataset 2 was acquired mostly with two smartphone models. The shift between smartphone models is also an important aspect for the use of these models in the field, since growers and technicians use different devices}. Neither domain-specific pre-training nor large-scale natural-image pre-training removes this shift.

The pronounced difference in Week 1 suggests that early phenological stages introduce visual structures that are insufficiently represented in Dataset 1, which results in a large distribution mismatch for both models. As the season progresses, the MMD to Dataset 1 becomes smaller, although even the most aligned weeks remain almost perfectly separable from Dataset 1.

These findings carry practical implications for the deployment of automated grapevine variety classification systems. The performance drop on Dataset 2 indicates that current models cannot be applied reliably to new acquisition years without additional adaptation. Systems intended for long-term and real-world use will require increased dataset diversity that includes multi-year imagery, as well as techniques for reducing distribution discrepancies. Incremental retraining as new data becomes available can also help maintain model calibration over time. While Dataset 3+ pre-training improves in-season accuracy and label efficiency, it must be complemented with extensive cross-year labelled data to achieve the level of robustness required for operational scenarios. Early-season samples in particular require additional attention, since their reduced morphological differentiation poses a consistent challenge that is not resolved by the pre-training strategies evaluated here.

\subsection{Comparison with other SSL Methods}

ViT-S pre-trained with MAE, DINO and Barlow Twins exhibits distinct behaviours. DINO reaches the best performance early in training, which makes it attractive when rapid model development is required. However, its performance stabilises around epoch five hundred and then begins to decrease. This trend is reflected in the effective rank, which also drops after the plateau and indicates that the model gradually loses feature diversity as training progresses. Although DINO is the fastest to converge, each epoch requires more than twice the GPU time of MAE, which affects its suitability for large-scale or resource-limited pre-training.

Barlow Twins achieves the highest effective rank throughout the experiments, which is expected from its redundancy-reduction objective, since its representations spread variance over more dimensions. This did not translate into better fine-tuning performance, since Barlow Twins obtained the lowest F1 score of the three methods. In our experiments, a higher effective rank was therefore not a reliable predictor of downstream accuracy across methods.

MAE presents the lowest effective rank, and although long training slightly improves its value, the overall shape of the curve remains consistent. This behaviour suggests a relation between effective rank and the exploration phase of training, which is strongly influenced by the learning rate schedule. The reduced effective rank is expected for reconstruction-based approaches, where the model focuses on reconstructing masked patches rather than contrasting different samples. This dimensional collapse \gc{may be} a disadvantage in a fine-grained domain such as grapevine variety classification, where the subtle differences between classes require a large number of features, and reducing it is an important direction for improving MAE in this domain. MAE needed a longer schedule than DINO to reach a similar F1 score, but each epoch was cheaper.

SimSiam does not converge and produces collapsed representations in our setting. In the agricultural domain, \citet{david_dong_self-supervised_2023} reported similar issues with SimSiam's convergence.

The UMAP projections (\Cref{fig:met_umap}) are consistent with these observations, since clusters corresponding to different grapevine varieties are visible for all methods before fine-tuning, and the projections become more similar after fine-tuning, as expected when supervised fine-tuning aligns the representations with the classification task.

From a practical viewpoint, MAE is attractive when long pre-training on a single GPU is feasible, because of its low cost per epoch and modest batch size requirements, while DINO reached a good result in fewer epochs at a higher cost per epoch. In addition, MAE is known to perform particularly well when the whole model is fine-tuned \citep{he_masked_2021}, which is the protocol used in this study.

Finally, to the best of our knowledge, only the study by  \citet{yousra_self-supervised_2023} has compared MAE with another self-supervised learning method in an agricultural context. Their results showed that SimCLR  \citep{chen_simple_2020} outperformed MAE in identifying pigs, although specific pre-training details were not provided. This highlights the lack of comprehensive comparisons of SSL strategies in agriculture and reinforces the need for further investigation into how different methods behave across diverse tasks and imaging conditions.

Overall, the choice of method should consider the balance between convergence speed, peak performance, computational cost and the characteristics of the target application. Within a single method, the effective rank followed the fine-tuning performance during training (for example, for DINO), which suggests that monitoring it may help to choose training schedules; across methods, however, it did not predict performance.

\subsection{Comparison with Similar Studies}

Among studies applying deep learning techniques to classify at least 10 grapevine varieties using RGB images captured in the field \citep{de_nart_vine_2024, terzi_automatic_2024, magalhaes_toward_2023, liu_development_2021, carneiro_evaluating_2023, carneiro_2023, carneiro_grapevine_2022}, reported performance metrics (accuracy or F1-Score) range from 0.8900 to 0.9991. A comparison among the studies can be seen in \gc{\Cref{tab:sim_studies}}.

\citet{de_nart_vine_2024} achieved 0.9970 accuracy using an Inception V3 to classify 27 varieties with images captured both in the field and in controlled environments across two seasons. Unlike this study, they excluded images of leaves that were deformed, overly mature, or too young.  \citet{magalhaes_toward_2023} classified 26 varieties using images acquired in a controlled environment during a single season, focusing on specific plant positions, simplifying classification. \citet{liu_development_2021} reported 0.9991 of accuracy with GoogleLeNet for 21 varieties using images captured in the field, although the acquisition period was unspecified.

\citet{terzi_automatic_2024} employed a dataset of 50 varieties, consisting of leaf and bunch images taken in the field. However, their classification was conducted using subsets of 10 or 11 varieties to mitigate memorisation effects. Meanwhile, \citet{carneiro_analyzing_2022, carneiro_grapevine_2022, carneiro_evaluating_2023, carneiro_2023} conducted earlier experiments with subsets of Dataset 1 containing 12 or 14 varieties.

To the best of our knowledge, this study is the first to classify grapevine varieties using a dataset comprising over 40 varieties simultaneously, collected across five growing seasons (2020 to 2024). The dataset includes deformed images, various plant positions and plants of different ages. The scores obtained here are lower than those reported in previous studies, which used fewer varieties and data acquired in a single place, on a few days of the same season or in controlled environments. Since the datasets, numbers of classes and metrics differ, the values in \Cref{tab:sim_studies} are not directly comparable, but they show how much harder the present benchmark is.

\begin{table}
	\centering
	\footnotesize
	\caption{Comparison of studies applying deep learning techniques for grapevine variety classification using datasets with a minimum of 10 classes. Values come from different datasets and metrics and are not directly comparable.}
	\label{tab:sim_studies}
	\begin{tabularx}{\linewidth}{l c c c c c}
		\toprule
		\textbf{Study} & \textbf{Varieties} & \textbf{Image Source} & \textbf{Seasons} & \textbf{Model} & \textbf{Performance} \\
		\midrule
		\citet{de_nart_vine_2024}          & 27  & Field + Controlled & 2              & Inception V3      & 0.9970 (Acc.) \\
		\citet{magalhaes_toward_2023}      & 26  & Controlled         & 1              & MobileNetV2       & 0.9475 (F1)   \\
		\citet{liu_development_2021}       & 21  & Field              & Not specified  & GoogleLeNet       & 0.9991 (Acc.) \\
		\citet{terzi_automatic_2024}       & 50 (10 to 11 used)       & Field           & Not specified  & Handcrafted       & 0.9720 (Acc.) \\
		\citet{carneiro_analyzing_2022}    & 12  & Field              & 1  & Xception          & 0.9200 (F1)   \\
		\citet{carneiro_evaluating_2023}   & 14  & Field              & 2  & EfficientNetV2S   & 0.8900 (F1)   \\
		\citet{carneiro_grapevine_2022}    & 12  & Field              & 1  & ViT-B             & 0.9600 (F1)   \\
		\textbf{This study}                & \textbf{43} & \textbf{Field} & \textbf{5} & \textbf{ViT-B} & \textbf{0.7956 (F1)} \\
		\bottomrule
	\end{tabularx}
\end{table}

\subsection{Implications for Precision Agriculture}

The findings of this study have implications for the application of MAE-based models in precision agriculture. Models pre-trained with MAE on grapevine-specific data provide a label-efficient starting point for variety identification, but their use in new seasons still requires labelled data from those seasons. Accurate classification of grapevine varieties is critical for vineyard management and quality control in wine production, as understanding variety distribution within vineyards allows for practices tailored to specific requirements \citep{Rankine2017lnfluenceOG, Fanzone2012PhenolicCO, SanchoGaln2020IdentificationAC}.

The potential to extend the use of MAEs to other agricultural domains is considerable. Self-supervised models pre-trained on crop-specific datasets could enable applications such as disease detection, growth stage monitoring and yield estimation, where detailed pattern recognition is essential. Furthermore, integrating MAEs with other self-supervised approaches, such as contrastive learning methods \citep{chen_simple_2020, chen_improved_2020}, could enhance feature extraction capabilities, particularly in scenarios with high intra-class variability.

The observed performance drop during early phenological stages and the distribution shift between seasons indicate that practical deployment must account for temporal variability in leaf morphology and environmental conditions. Reliable operation in vineyard settings requires training data that span multiple years and growth stages, or the use of adaptation strategies that compensate for distribution differences. Early season monitoring tasks may be particularly affected, since reduced morphological differentiation limits classification accuracy during this period. Models pre-trained on domain-specific datasets, such as Dataset 3+, offer advantages in capturing class-relevant features, but extensive multi-season data and adaptation mechanisms remain necessary to ensure robust performance for operational use.

\subsection{Limitations and Future Directions}

The results show that, for identifying grapevine varieties through images, MAE pre-training on a relatively small set of domain-specific images is competitive with, and in our setting slightly better than, the transfer from the public MAE IN1K checkpoint. However, the study has some limitations. Firstly, all the results come from single runs, so small differences (for example, 0.7956 against 0.7851 after merging, or 0.4677 against 0.4747 on Dataset 2) should be considered with caution. Secondly, the supervised baselines were fine-tuned with a different recipe from the MAE models. \gc{Thirdly, the images of Dataset 1 were assigned to the subsets at random, so images of the same plant can appear in the training and test subsets, which may lead to optimistic scores on the Dataset 1 test subset.}

\gc{In addition to these limitations of the experimental design, three characteristics of grapevine variety identification make the task itself difficult.} Firstly, there are between 5,000 and 8,000 grapevine varieties spread across the globe \citep{robinson_wine_2013, schneider_verifying_2001}, necessitating a mechanism to rapidly incorporate new varieties into the classifier’s scope. Secondly, grapevine variety identification can be viewed as a long-tailed data distribution problem, where some varieties are more common than others. This issue is particularly pronounced in regions with a large number of local and indigenous varieties, such as Italy or Portugal, where many varieties may only be known to a few producers. This makes it challenging to gather representative samples of these rare varieties. Thirdly, as the goal is to classify different varieties within the same species, the task is also a fine-grained classification problem. Therefore, there is still considerable room for improvement in this study.

To address these challenges, future work could leverage the pre-trained weights obtained from this exploration as initial weights for zero-shot learning \citep{chen_knowledge-aware_2021} and incremental learning \citep{liu_incremental_2023} approaches to handle the addition of new varieties. Additionally, few-shot learning \citep{parnami_learning_2022} can be applied to manage naturally under-represented classes, and the architecture could be adapted \citep{he_transfg_2021}, or feature enhancement methods \citep{pan_ssfe-net_2023} could be used to improve performance, considering the fine-grained classification nature of the task.

\section{Conclusions}
\label{sec:conclusions}
This study evaluated the use of masked autoencoders as an alternative to IN1K transfer learning in the identification of 43 grapevine varieties using field-acquired leaf images. The ViT-B model pre-trained with MAE on an unlabelled dataset of 54,571 images achieved the best performance on Dataset 1 (F1-Score of 0.7776 when fine-tuned on Dataset 1 and 0.7956 when fine-tuned on Datasets 1 and 2), 0.0282 above the public MAE IN1K checkpoint fine-tuned with the same recipe. A benchmark dataset of 43 grapevine varieties acquired over five growing seasons was also presented, which is, to the best of our knowledge, the largest public benchmark for this task in number of varieties. This benchmark includes 5 of the 13 most widely planted grape varieties globally.

The evaluation indicated that models benefit from long pre-training (between 2000 and 2500 epochs), and models pre-trained on datasets that align closely with downstream tasks perform better when trained in low-data regimes. Additionally, it was found that random cropping leads to better performance compared to stronger augmentations like colour jitter or Gaussian blur. The results also show that varying the mask ratio between 0.50 and 0.75 slightly affects the final classification performance.

The generalisation to a new season remains the main challenge. The models trained on Dataset 1 lost about 40\% of their F1 score on the 2024 images, regardless of the pre-training, and \gc{the feature-space analysis of both domain-specific and IN1K pre-trained encoders showed that each acquisition season is distinct, with the 2024 images of the early phenological stages the most distant from the training data.} An analysis of the classifier’s performance across three different weeks in the 2024 season was conducted. The results indicate that the classifier’s worst performance occurred towards the end of May, when plants are in leaf development and inflorescence emergence. Multi-year, seasonally diverse labelled data therefore remain necessary for reliable deployment.

For future work, repeated runs will be used to estimate the variability of the results, and images from new seasons and vineyards will be acquired to improve the generalisation of the models. \gc{The effects of the season and of the acquisition device on the distribution shift will also be separated, and a plant-disjoint split of the benchmark will be prepared.} In addition, the pre-trained weights obtained from this study can be used to implement incremental learning and zero-shot learning approaches, which would allow for the rapid inclusion of new varieties. Few-shot learning methods could also be applied to address the naturally under-represented varieties. Moreover, architecture modifications and feature enhancement techniques could be employed to further improve performance, as identifying grapevine varieties remains a fine-grained classification task.

\section*{Data and Code Availability}
Datasets 1 and 2 are publicly available at \url{https://zenodo.org/records/15160944}. Datasets 3 and 3+ combine images from third-party sources whose licences do not all allow redistribution; to enable their reconstruction, we provide the list of sources and image identifiers or URLs.

\section*{Financial support}
The authors would like to acknowledge the Portuguese Foundation for Science and Technology (FCT) for financial support through national funds to the projects UIDB/04033/2020 (\url{https://doi.org/10.54499/UIDB/04033/2020}) (accessed on 1 December 2024) as well as through Gabriel Carneiro's doctoral scholarship (\url{https://doi.org/10.54499/PRT/BD/154883/2023}), and also the Vine \& Wine Portugal Project, co-financed by the Recovery and Resilience Plan (RRP) and the European NextGeneration EU Funds, within the scope of the Mobilizing Agendas for Reindustrialization, under ref. C644866286-00000011.

\section*{Competing interests}
None.

\section*{Author Contributions}
G.A.C. designed the study, developed the methodology, performed the experiments and analysis, and wrote the original draft. T.J.A., A.C., and P.R. contributed to the study design, supervised the work, and critically revised the manuscript. J.S. contributed to the study design and methodology, supervised the project, handled project administration, secured funding, and critically revised the manuscript.

All authors reviewed and approved the final manuscript.

\section*{Use of AI Tools}
Claude (Anthropic) was used during the preparation of this manuscript to assist with grammar and language correction. The tool was used solely for improving linguistic clarity and did not contribute to the development of the scientific content, analysis, interpretation, or conclusions of the work. The authors take full responsibility for the content and integrity of the manuscript.

\bibliographystyle{plainnat}
\bibliography{bib}

\clearpage
\appendix
\section{Per-class results}\crefalias{section}{appendix}\label{app:perclass}%

This appendix contains the per-class tables referred to in \gc{\Cref{sec:results,sec:seasonal}}. Class codes and the number of images per class in each subset of Dataset 1 are given in \Cref{fig:dataset1_samples}.

\begingroup
\singlespacing
\small
\setlength{\tabcolsep}{3pt}
\renewcommand{\arraystretch}{0.9}

\begin{longtable}{@{}ccccccc@{}}
	\caption{Per-class results on the Dataset 1 test subset for ViT-B pre-trained with MAE on Dataset 3+ and IN1K, both fine-tuned on Dataset 1. Class codes are defined in \Cref{fig:dataset1_samples}, which also gives the number of test images per class.}
	\label{tab:app_complete_results} \\

	\toprule
	\multirow{2}{*}{\textbf{Class}} & \multicolumn{3}{c}{\textbf{Dataset 3+}} & \multicolumn{3}{c}{\textbf{IN1K}} \\
	\cmidrule(l){2-7}
	& \textit{Precision} $\uparrow$ & \textit{Recall} $\uparrow$ & \textit{F1-Score} $\uparrow$ & \textit{Precision} $\uparrow$ & \textit{Recall} $\uparrow$ & \textit{F1-Score} $\uparrow$ \\
	\midrule
	\endfirsthead

	\toprule
	\multirow{2}{*}{\textbf{Class}} & \multicolumn{3}{c}{\textbf{Dataset 3+}} & \multicolumn{3}{c}{\textbf{IN1K}} \\
	\cmidrule(l){2-7}
	& \textit{Precision} $\uparrow$ & \textit{Recall} $\uparrow$ & \textit{F1-Score} $\uparrow$ & \textit{Precision} $\uparrow$ & \textit{Recall} $\uparrow$ & \textit{F1-Score} $\uparrow$ \\
	\midrule
	\endhead

	\midrule
	\multicolumn{7}{r}{\textit{Continued on next page}}
	\endfoot

	\bottomrule
	\endlastfoot

	AC & 1.0000 & 1.0000 & 1.0000 & 0.8889 & 1.0000 & 0.9412 \\
	AV & 0.9433 & 0.9779 & 0.9603 & 0.9097 & 0.9632 & 0.9357 \\
	AT & 0.6119 & 0.8200 & 0.7009 & 0.7255 & 0.7400 & 0.7327 \\
	BT & 0.8750 & 0.7368 & 0.8000 & 0.6667 & 0.8421 & 0.7442 \\
	BA & 0.7442 & 0.9412 & 0.8312 & 0.6977 & 0.8824 & 0.7792 \\
	CF & 0.7222 & 0.6842 & 0.7027 & 0.6818 & 0.7895 & 0.7317 \\
	CS & 0.9444 & 0.8947 & 0.9189 & 0.8571 & 0.9474 & 0.9000 \\
	CN & 0.8235 & 0.7368 & 0.7778 & 0.7778 & 0.7368 & 0.7568 \\
	CC & 0.7288 & 0.8958 & 0.8037 & 0.6780 & 0.8333 & 0.7477 \\
	CD & 0.8824 & 0.7895 & 0.8333 & 0.8125 & 0.6842 & 0.7429 \\
	CG & 0.8980 & 0.8165 & 0.8553 & 0.8836 & 0.8266 & 0.8542 \\
	CR & 0.6907 & 0.8701 & 0.7701 & 0.6465 & 0.8312 & 0.7273 \\
	CT & 0.8095 & 0.9444 & 0.8718 & 0.7000 & 0.7778 & 0.7368 \\
	DB & 0.6667 & 0.7632 & 0.7117 & 0.6556 & 0.7763 & 0.7108 \\
	DT & 0.8000 & 0.6316 & 0.7059 & 0.7778 & 0.7368 & 0.7568 \\
	EC & 0.4848 & 0.5333 & 0.5079 & 0.5278 & 0.6333 & 0.5758 \\
	FP & 0.6875 & 0.6875 & 0.6875 & 0.5417 & 0.8125 & 0.6500 \\
	FG & 0.7231 & 0.7833 & 0.7520 & 0.6615 & 0.7167 & 0.6880 \\
	GM & 0.8333 & 0.8333 & 0.8333 & 0.9231 & 0.6667 & 0.7742 \\
	GV & 0.4898 & 0.7059 & 0.5783 & 0.4490 & 0.6471 & 0.5301 \\
	MC & 0.5500 & 0.5789 & 0.5641 & 0.3846 & 0.5263 & 0.4444 \\
	MF & 0.6701 & 0.6989 & 0.6842 & 0.6778 & 0.6559 & 0.6667 \\
	MP & 0.7188 & 0.8214 & 0.7667 & 0.6471 & 0.7857 & 0.7097 \\
	MR & 0.4828 & 0.5385 & 0.5091 & 0.5455 & 0.4615 & 0.5000 \\
	ML & 0.8750 & 0.7368 & 0.8000 & 1.0000 & 0.7895 & 0.8824 \\
	MG & 0.9482 & 0.8491 & 0.8959 & 0.9412 & 0.8295 & 0.8818 \\
	MX & 0.2727 & 0.6316 & 0.3810 & 0.1940 & 0.6842 & 0.3023 \\
	MT & 0.8065 & 0.8621 & 0.8333 & 0.7353 & 0.8621 & 0.7937 \\
	PB & 0.6522 & 0.7895 & 0.7143 & 0.6538 & 0.8947 & 0.7556 \\
	RB & 0.7829 & 0.8860 & 0.8313 & 0.8276 & 0.8421 & 0.8348 \\
	RF & 0.8000 & 0.8421 & 0.8205 & 0.8824 & 0.7895 & 0.8333 \\
	SM & 0.5862 & 0.6711 & 0.6258 & 0.6269 & 0.5526 & 0.5874 \\
	SB & 0.7368 & 0.7778 & 0.7568 & 0.6471 & 0.6111 & 0.6286 \\
	SS & 0.9583 & 0.9787 & 0.9684 & 0.8824 & 0.9574 & 0.9184 \\
	TA & 0.7681 & 0.8983 & 0.8281 & 0.7941 & 0.9153 & 0.8504 \\
	TB & 0.8649 & 0.9275 & 0.8951 & 0.8148 & 0.9565 & 0.8800 \\
	TM & 0.7222 & 0.8125 & 0.7647 & 0.6316 & 0.7500 & 0.6857 \\
	TF & 0.7222 & 0.6842 & 0.7027 & 0.6250 & 0.5263 & 0.5714 \\
	TR & 0.9428 & 0.9276 & 0.9351 & 0.9455 & 0.9303 & 0.9378 \\
	TC & 0.8750 & 0.9655 & 0.9180 & 0.9333 & 0.9655 & 0.9492 \\
	TS & 0.7692 & 0.9091 & 0.8333 & 0.7949 & 0.9394 & 0.8611 \\
	TN & 0.9807 & 0.9568 & 0.9686 & 0.9864 & 0.9530 & 0.9694 \\
	VO & 0.8235 & 0.8485 & 0.8358 & 0.7429 & 0.7879 & 0.7647 \\

	\midrule
	Macro Average & 0.7597 & 0.8056 & 0.7776 & 0.7297 & 0.7863 & 0.7494 \\

\end{longtable}
\endgroup

\begingroup
\singlespacing
\small
\setlength{\tabcolsep}{3pt}
\renewcommand{\arraystretch}{0.9}

\begin{longtable}{@{}ccccc@{}}
	\caption{Per-class domain shift between Datasets 1 and 2, using features from ViT-B pre-trained with Dataset 3+ and IN1K, measured by held-out classifier AUC (five-fold cross-validation) and MMD (one bandwidth per encoder, as in \gc{\Cref{tab:domain_shift_weeks}}), on the validation subsets. $^{\dagger}$ marks shifts that are not significant ($p \geq 0.05$, MMD permutation test), and all the other classes have $p < 0.05$.}
	\label{tab:domain_shift} \\

	\toprule
	& \multicolumn{2}{l}{\textbf{Dataset 3+}} & \multicolumn{2}{l}{\textbf{IN1K}} \\
	\midrule
	\textbf{Class} & \textit{AUC} $\downarrow$ & \textit{MMD} $\downarrow$ & \textit{AUC} $\downarrow$ & \textit{MMD} $\downarrow$ \\
	\midrule
	\endfirsthead

	\toprule
	& \multicolumn{2}{l}{\textbf{Dataset 3+}} & \multicolumn{2}{l}{\textbf{IN1K}} \\
	\midrule
	\textbf{Class} & \textit{AUC} $\downarrow$ & \textit{MMD} $\downarrow$ & \textit{AUC} $\downarrow$ & \textit{MMD} $\downarrow$ \\
	\midrule
	\endhead

	\midrule
	\multicolumn{5}{r}{\textit{Continued on next page}}
	\endfoot

	\bottomrule
	\endlastfoot

	Alicante & 0.9746 & 0.3076 & 0.9587 & 0.2508 \\
	Alvarelh\~ao & 0.9725 & 0.2236 & 0.9081 & 0.1196 \\
	Arinto & 0.9484 & 0.2514 & 0.9371 & 0.2663 \\
	Bastardo & 0.9035 & 0.1873 & 0.7807 & 0.2570 \\
	Boal & 0.9271 & 0.1789 & 0.9527 & 0.1792 \\
	Cabernet Franc & 0.9444 & 0.3051 & 0.9386 & 0.3813 \\
	Cabernet Sauvignon & 0.8538 & 0.1041$^{\dagger}$ & 0.6842 & 0.1045$^{\dagger}$ \\
	Carignan Noir & 0.7593 & 0.1957 & 0.6944 & 0.2366 \\
	Cercial & 0.9781 & 0.2711 & 0.9775 & 0.2627 \\
	Chardonnay & 0.8056 & 0.1277$^{\dagger}$ & 0.7778 & 0.1439$^{\dagger}$ \\
	C\'odega & 0.9844 & 0.1914 & 0.9797 & 0.1983 \\
	C\'odega do Larinho & 0.9886 & 0.3087 & 0.9561 & 0.3145 \\
	Cornifesto & 0.9074 & 0.1864 & 0.8259 & 0.1424 \\
	Donzelinho Branco & 0.9852 & 0.2534 & 0.9652 & 0.2971 \\
	Donzelinho Tinto & 0.9286 & 0.1774 & 0.9286 & 0.2274 \\
	Esgana C\~ao & 0.9258 & 0.1547 & 0.8785 & 0.2021 \\
	Fern\~ao Pires & 1.0000 & 0.2937 & 0.9909 & 0.3411 \\
	Folgas\~ao & 0.9364 & 0.2194 & 0.9066 & 0.2167 \\
	Gamay & 0.8977 & 0.1894 & 0.8743 & 0.2736 \\
	Gouveio & 0.8580 & 0.1584 & 0.8068 & 0.1623 \\
	Malvasia Corada & 0.9977 & 0.2866 & 0.9519 & 0.3001 \\
	Malvasia Fina & 0.9865 & 0.2399 & 0.9609 & 0.2302 \\
	Malvasia Preta & 0.9209 & 0.2126 & 0.8750 & 0.2541 \\
	Malvasia Rei & 1.0000 & 0.3189 & 1.0000 & 0.3089 \\
	Merlot & 0.9678 & 0.2503 & 0.8830 & 0.3271 \\
	Moscatel Galego & 0.9646 & 0.2003 & 0.9563 & 0.1680 \\
	Moscatel Galego Roxo & 0.9683 & 0.2786 & 0.9524 & 0.2632 \\
	Mourisco Tinto & 0.8383 & 0.1588 & 0.8139 & 0.1918 \\
	Pinot Blanc & 0.9037 & 0.2993 & 0.8815 & 0.3830 \\
	Rabigato & 0.9227 & 0.1760 & 0.9545 & 0.1501 \\
	Rufete & 0.8509 & 0.1613 & 0.6725 & 0.1458 \\
	Samarrinho & 0.9253 & 0.2252 & 0.8914 & 0.2467 \\
	Sauvignon Blanc & 0.8819 & 0.2370 & 0.8889 & 0.3734 \\
	Sous\~ao & 0.9793 & 0.3015 & 0.9617 & 0.2105 \\
	Tinta Amarela & 0.9670 & 0.2087 & 0.9483 & 0.2497 \\
	Tinta Barroca & 0.9729 & 0.1822 & 0.9417 & 0.2020 \\
	Tinta F\^emea & 0.9739 & 0.2594 & 0.9478 & 0.2241 \\
	Tinta Francisca & 0.8556 & 0.1914 & 0.9074 & 0.2564 \\
	Tinta Roriz & 0.9844 & 0.2460 & 0.9453 & 0.2132 \\
	Tinto C\~ao & 0.9024 & 0.2131 & 0.8357 & 0.1818 \\
	Touriga Francesa & 0.8532 & 0.1864 & 0.8911 & 0.1905 \\
	Touriga Nacional & 0.9708 & 0.1869 & 0.8609 & 0.1390 \\
	Viosinho & 0.8854 & 0.1529 & 0.8116 & 0.1293$^{\dagger}$ \\

	\midrule
	Average & 0.9291 & 0.2200 & 0.8943 & 0.2306 \\

\end{longtable}
\endgroup

\end{document}